%% file: main.tex
\PassOptionsToPackage{dvipsnames}{xcolor}

\documentclass{article} 

\input{preamble}

\begin{document}

\twocolumn[
\icmltitle{\approachbf: Efficient Evolutionary Merging \\ on Consumer-grade GPUs}


\icmlsetsymbol{equal}{*}

\begin{icmlauthorlist}
    \icmlauthor{Tommaso Mencattini}{equal,EPFL}
    \icmlauthor{Adrian Robert Minut}{equal,Sapienza University of Rome}
    \icmlauthor{Donato Crisostomi}{Sapienza University of Rome}
    \icmlauthor{Andrea Santilli}{Sapienza University of Rome}
    \icmlauthor{Emanuele Rodolà}{Sapienza University of Rome}
\end{icmlauthorlist}

\icmlaffiliation{EPFL}{Ecole Polytechnique Fédérale de Lausanne}
\icmlaffiliation{Sapienza University of Rome}{Sapienza University of Rome}

\icmlcorrespondingauthor{Tommaso Mencattini}{tommaso.mencattini@epfl.ch}

\icmlkeywords{Model Merging, Evolutionary Algorithms, Efficient Methods for Machine Learning, LLMs, Multilingual Models}

\vskip 0.3in
]



\printAffiliationsAndNotice{\icmlEqualContribution} % otherwise use the standard text.

\begin{abstract}
    Evolutionary model merging enables the creation of high-performing multi-task models but remains computationally prohibitive for consumer hardware.
    We introduce \approachns, an efficient framework that makes evolutionary merging feasible on a single GPU by reducing fitness computation costs 50× while preserving performance. \approachns{} achieves this by \textbf{E}xtracting a reduced dataset for evaluation, \textbf{E}stimating model abilities using Item Response Theory (IRT), and \textbf{E}volving optimal merges via IRT-based performance estimators. Our method enables state-of-the-art multilingual and cross-lingual merging, transferring knowledge across languages with significantly lower computational overhead.
    We provide theoretical guarantees and an open-source library\footnote{\href{https://github.com/tommasomncttn/merge3}{github.com/tommasomncttn/merge3}}, democratizing high-quality model merging.
\end{abstract}

\section{Introduction}
\input{1_Introduction/content}

\section{Related Work}
\input{2_Background/content}

\section{\approachbf}
\input{3_Approach/content}

\section{Experiments}
\input{4_Experiments/content}

\section{Theoretical Analysis}
\input{5_Theoretical_analysis/content}

\section{Conclusions}
\input{7_Conclusions/content}

\subsubsection*{Acknowledgments}
We acknowledge ISCRA for awarding this project access to the LEONARDO supercomputer, owned by the EuroHPC Joint Undertaking, hosted by CINECA (Italy).

\bibliography{references,FMP}
\bibliographystyle{icml2025}

\newpage

\appendix

\section{Mergenetic}
\input{E_Library/content}

\section{Additional Details}
\input{A_Details/content}

\section{Additional Experiments}
\input{B_Additional_Experiments/content}

\section{Mathematical proofs}\label{app:proofs}
\input{D_Proofs/content}

\end{document}

%% file: preamble.tex
\usepackage{times}
\usepackage{graphicx}
\usepackage{booktabs}
\usepackage{xcolor}
\usepackage{amsmath, amsthm, amssymb, amsfonts}
\usepackage{thmtools}
\usepackage{graphicx}
\usepackage{setspace}
\usepackage{float}
\usepackage{hyperref}
\usepackage[utf8]{inputenc}
\usepackage[english]{babel}
\usepackage{framed}
\usepackage{cleveref}
\usepackage[inline]{enumitem}
\usepackage{hyperref}
\usepackage{url}
\usepackage{wrapfig}
\usepackage{subcaption}
\usepackage[accepted]{icml2025}
\usepackage{comment}
\usepackage{multirow}

\usepackage{bm}
\newcommand{\ones}{\bm{1}}
\newtheorem{theorem}{Theorem}
 
\newtheorem{assumption}{Assumption}
\declaretheoremstyle[
    headfont=\bfseries, 
    bodyfont=\normalfont, 
    spaceabove=6pt, 
    spacebelow=6pt
]{defsty}
\declaretheorem[style=defsty,numberlike=theorem]{definition}

\declaretheoremstyle[
    headfont=\bfseries, 
    bodyfont=\normalfont, 
    spaceabove=6pt, 
    spacebelow=6pt
]{defsty}

\declaretheoremstyle[name=Proposition,]{prosty}
\declaretheorem[style=prosty,numberlike=theorem]{proposition}

\declaretheoremstyle[name=conjecture,]{prcpsty}

\input{math_commands.tex}

\icmltitlerunning{MERG$E^3$: Fast Evolutionary Merging via \\ Data Extraction and Ability Estimation}

\newcommand{\approach}[0]{MERGE$^3$~}
\newcommand{\approachns}[0]{MERGE$^3$}
\newcommand{\approachbf}[0]{$\mathbf{\text{MERGE}^3}$}
\newcommand{\model}[1]{\texttt{#1}}
\newcommand{\dataset}[1]{\texttt{#1}}
\newcommand{\method}[1]{\texttt{#1}}

\newcommand{\improvUP}[1]{\color{PineGreen}($\uparrow$#1)}

\crefname{conjecture}{Conjecture}{Conjectures}%

\hypersetup{
    colorlinks,
    linkcolor={red!50!black},
    citecolor={blue!50!black},
    urlcolor={blue!80!black}
}

\usepackage[textsize=tiny]{todonotes}

\icmltitlerunning{Efficient Evolutionary Merging on Consumer-grade GPUs}

\newcommand{\Call}[2]{\textsc{#1}\!\bigl(#2\bigr)}
\newcommand{\Comment}[1]{\hfill \textcolor{gray}{\# #1}}

\newcommand{\mpirt}{%
  \ifmmode
    \text{\textsc{mp-IRT}}%
  \else
    \textsc{mp-IRT}%
  \fi
}

\newcommand{\pirt}{%
  \ifmmode
    \text{\textsc{p-IRT}}%
  \else
    \textsc{p-IRT}%
  \fi
}

\newcommand{\gpirt}{%
  \ifmmode
    \text{\textsc{gp-IRT}}%
  \else
    \textsc{gp-IRT}%
  \fi
}

\newcommand{\gmpirt}{%
  \ifmmode
    \text{\textsc{gmp-IRT}}%
  \else
    \textsc{gmp-IRT}%
  \fi
}

%% file: math_commands.tex
\usepackage{amsmath,amsfonts,bm}

\def\eqref#1{equation~\ref{#1}}
\def\1{\bm{1}}

\DeclareMathAlphabet{\mathsfit}{\encodingdefault}{\sfdefault}{m}{sl}
\SetMathAlphabet{\mathsfit}{bold}{\encodingdefault}{\sfdefault}{bx}{n}

%% file: 1_Introduction/content.tex
%!Tex root=../main.tex

Model merging has become a powerful and accessible approach for developing new state-of-the-art models without the need for cluster-grade computing typically required for large model training \cite{yang2024model}.
Its key advantage lies in performing the merging process post-hoc directly in the parameters of pre-existing models (endpoint models), eliminating the need for training and significantly reducing the demand for expensive computational resources.

This approach has significantly broadened access to the field, with ML practitioners producing competitive models out of existing ones on standard consumer GPUs\footnote{At the time of writing, around 30\% of models on the Hugging Face Open LLM leaderboard are merged models.}\citep{task-vectors}.
\begin{figure}[t]
    \centering
    \includegraphics[width=.9\linewidth]{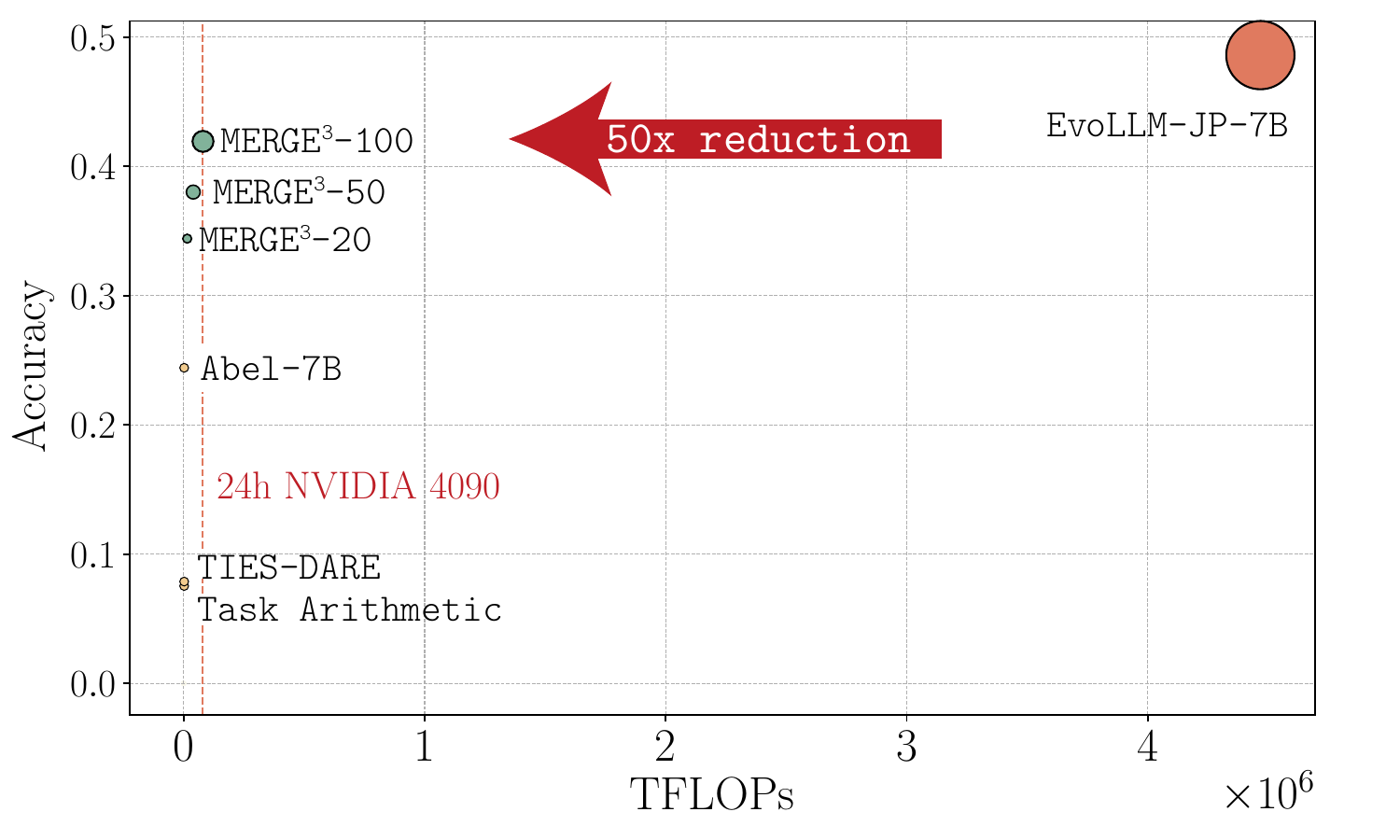}
    \caption{\textbf{Accuracy on Japanese \dataset{GSM8K} over fitness evaluation FLOPs.} \approach{} is competitive with a model evolved on the full dataset by only using a consumer-grade GPU and $2\%$ of the data (point size reflects data amount).}
    \label{fig:speed_accuracy}
\end{figure}
However, although computationally inexpensive, most of the existing approaches are quite rudimentary, require ad-hoc choices, and are usually based on ungrounded trial-and-error strategies for selecting the merge coefficients, which ultimately limits their downstream performance \citep{ties, yu2024language}.
On the other hand, recent work has shown that evolutionary merging can produce models of unprecedented quality by automating the hyperparameter search for merging coefficients \citep{sakana}.
While this technique can incorporate any standard merging method, such models are absent from public leaderboards likely due to a mismatch between the high computational demands of evolutionary merging and single-GPU setups typical of merging practitioners.
Indeed this computational cost is significantly high: computing the fitness function requires generating and evaluating answers for each dataset element, for each candidate in every evolutionary step. As shown in \Cref{fig:speed_accuracy}, the fitness computation alone in the 1,000-trial evolutionary merge from \citet{sakana} requires approximately $4\times10^6$ TFLOPs, with the full algorithm demanding largely over a month of continuous computation if run on a single NVIDIA 4090 (\S \ref{app:add-exp-4090}).
This renders evolutionary merging effectively out of reach for consumer hardware, risking to exclude the very user base it was meant to empower.

% APPROACH
In this paper, we address this challenge by introducing \approachns{}, an evolutionary merging framework that runs on a single consumer GPU with competitive results (see \cref{fig:speed_accuracy}). Unlike the competing approach, \approachns{} operates with just $0.077 \times 10^6$ TFLOPs, namely a {\bf 50-fold reduction}. This drastic decrease in computational cost makes it feasible on consumer hardware, freeing up FLOPs for further optimization or additional tasks.

\begin{figure}
    \centering
    \includegraphics[width=.99\linewidth]{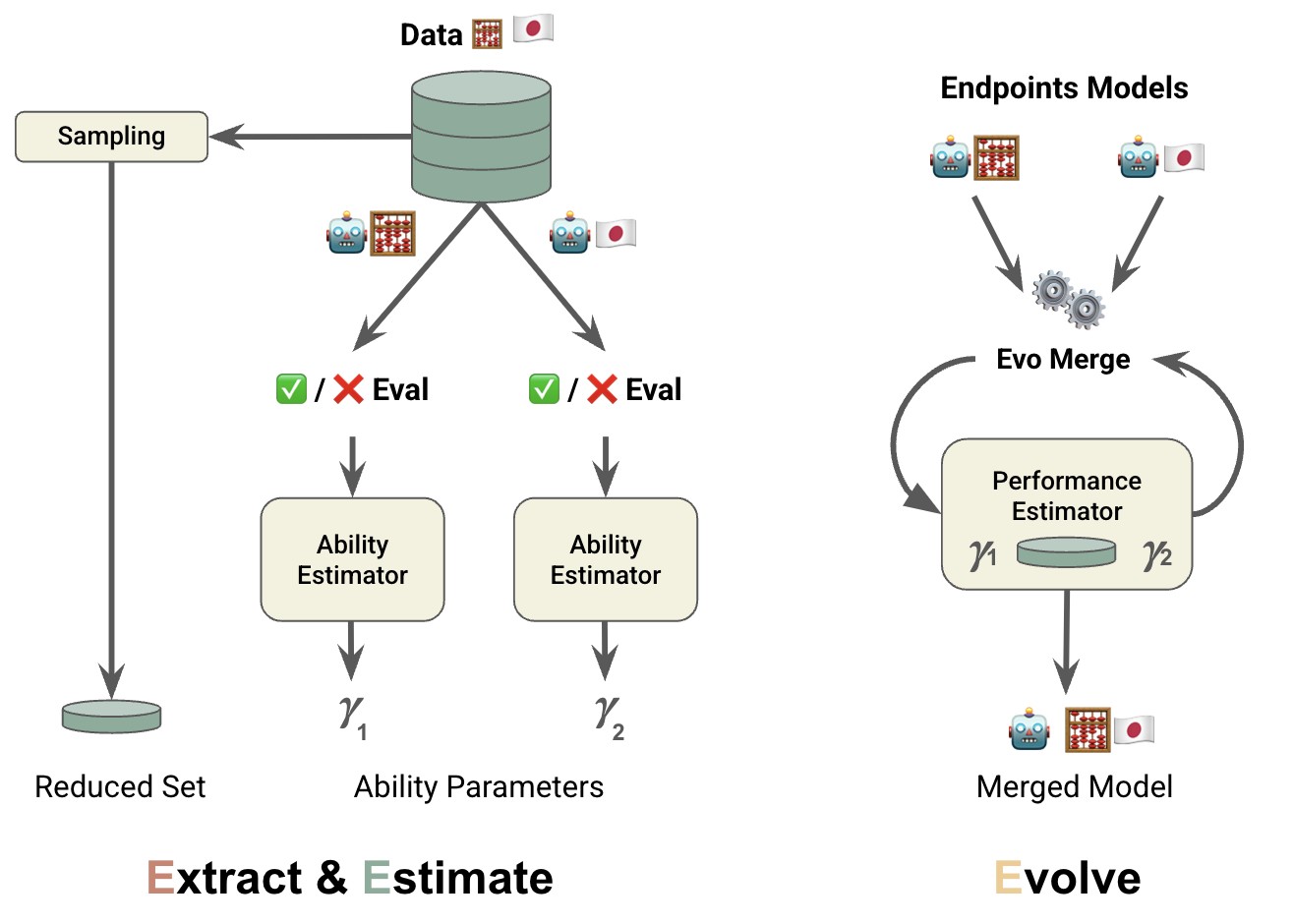}
    \caption{\textbf{\approachbf~for math + Japanese merging (\dataset{GSM8K})}.
    The method \textbf{Extracts} a reduced evolutionary dataset, \textbf{Estimates} ability parameters ($\gamma$) via Item Response Theory (IRT) based on their response correctness, and \textbf{Evolves} the endpoint models through iterative merging. Leveraging an IRT-based performance estimator, it approximates full-dataset fitness with reduced data, cutting fitness estimation costs while preserving full-dataset accuracy -- making evolutionary merging feasible on consumer GPUs.
    }
    \label{fig:teaser}
\end{figure}

Our approach starts by \textbf{E}xtracting a reduced subset of the fitness evaluation dataset, significantly alleviating the computational bottleneck of fitness computation (\cref{fig:teaser}). However, this reduction risks losing accuracy if the subset lacks diversity. To address this, we apply Item Response Theory (IRT) \cite{lord1968statistical}—a well-established statistical framework—to bridge the gap between reduced-dataset evaluations and full-dataset performance.
Specifically, we first \textbf{E}stimate the latent abilities of the endpoint models using IRT, ensuring the merged models accurately reflect their components' strengths. Then, we \textbf{E}volve the endpoint models with IRT-based performance estimators designed for model merging, assuming the merged model’s ability is a combination of those of the endpoint models. This approach significantly improves the efficiency and accuracy of fitness estimation, integrating merging-specific insights into performance estimation theory while maintaining high accuracy with reduced datasets.

Experimental results show that \approachns{} effectively transfers mathematical skills by merging a strong math model with three language-specific models, achieving 10–20\% higher accuracy than standard merging baselines in each language. Building on this, we evolve a single multilingual model by merging Italian, English, German, and Dutch models, outperforming individually fine-tuned models by up to 19\% on \dataset{ARC} \cite{ARC}, a widely used benchmark for reasoning.
Furthermore, \approachns{} achieves competitive accuracy on Japanese \dataset{GSM8K} \cite{gsm8k}, matching models evolved on full datasets while maintaining high efficiency, demonstrating that our evolutionary strategy preserves performance while drastically reducing computational costs.

To summarize, our contributions are fourfold:
\begin{itemize} 
    \item We introduce a novel, efficient evolutionary model merging framework leveraging Item Response Theory, making merging feasible on consumer hardware.
    \item We demonstrate its effectiveness in transferring skills across languages and synthesizing state-of-the-art multilingual models without standard training.
    \item We advance the theoretical foundations of performance estimation in model merging and provide formal guarantees for our proposed estimators.
    \item We release a modular library for evolutionary merging on consumer GPUs, alongside a suite of state-of-the-art models for several low-resource languages.
\end{itemize}

%% file: 2_Background/content.tex
% MODEL MERGING

\paragraph{Model Merging} has emerged as an efficient alternative to ensembling by integrating existing models without any additional training. One set of methods identifies neuron permutations that align the models into a shared optimization basin, allowing them to be merged through straightforward averaging \citep{git-rebasin, repair, zip-it, rebasin-implicit-sinkhorn, cycle-consistent}. Closer to our work, multi-task model merging focuses on the case where a single pre-trained model is fine-tuned for different tasks \citep{task-vectors, ties, yu2024language, matenamerging, wortsman2022model, davari2023model, wang2024localizing, zhou2024atm, gargiulo2025tasksingularvectorsreducing}. 
In this direction, several works address task interference by pruning or selectively combining parameters---e.g., TIES-merging~\citep{ties}, Model Breadcrumbs~\citep{davari2023model}, and DARE Merging~\citep{yu2024language}—or by optimizing merge coefficients \citep{yang2023adamerging}, introducing task-specific modules~\citep{yang2024representation}, and disentangling weights~\citep{ortiz2024task}. 

\paragraph{Evolutionary Algorithms.}
Evolutionary Algorithms are black-box optimization algorithms operating on a population of potential solutions by evolving them through generations with operators such as selection, mutation, recombination, and crossover \citep{6791438, petrowski2017evolutionary, dasgupta1997evolutionary}. 
Recent applications include neural architecture search~\citep{real2019regularized} and hyperparameter tuning~\citep{vincent2023improved}, where evolutionary methods efficiently navigate large design spaces without manual intervention. 
The fitness function is crucial, as it evaluates the quality of each solution, guiding the selection process by favoring higher-scoring (fitter) solutions for reproduction~\citep{ea}. Closest to our work, \citet{sakana} propose to apply evolutionary algorithms to optimize model merging recipes, eliminating the need for trial-and-error in combining parameters. In this context, the most obvious candidate for a fitness function is simply the performance of the resulting model over a held-out validation set. 

\paragraph{Item Response Theory.}
Item Response Theory (IRT) \citep{cai2016item, van2018handbook, brzezinska2020item, lord1968statistical} is a paradigm to design, analyze, and score responses to tests such as SAT or GRE \citep{an2014item, kingston1982feasibility, petersen1982using}. Based on the relationship between individuals' performances on a test item and the test takers' levels of performance on the corresponding required ability, IRT has recently spread from psychometrics to natural language processing. In this direction, \citet{lalor2016building} leverage IRT's latent dimensions to evaluate language models, while \citet{vania2021comparing} use it to analyze benchmark saturation in NLP evaluations. 
More relevant to our work, \citet{zhuang2023efficiently} and \citet{tinybenchmarks} employ IRT-driven adaptive testing to alleviate the computational burden of large-scale evaluations for large language models (LLMs). Although their focus is on LLM evaluation, which shares similarities with the efficient evaluation of fitness functions in model merging, our work builds on these approaches to design IRT-based estimators specifically tailored for model merging. Unlike prior applications of IRT, which are limited to LLM evaluations, our approach adapts the framework to address the unique challenges of evolutionary model merging, enabling efficient and accurate fitness estimation.

%% file: 3_Approach/content.tex
Our method \approach speeds up evolutionary model merging by reducing the computational cost of fitness evaluation. It achieves this by shrinking the fitness evaluation dataset and using IRT-based performance estimators to maintain full-dataset accuracy from subset evaluations. \Cref{fig:teaser} shows an overview of our method (\cref{alg:merge3}).

\subsection{Extract \& Estimate} \label{sec: extract_and_estimate}
Evaluating the fitness function involves generating and assessing answers for each data sample, repeated across all models in the population at every evolutionary step. Given the computational demands of evolutionary algorithms and LLMs, this process is highly intensive. To mitigate this, we reduce the dataset $D$ to a smaller subset $\bar{D} \subset D$ with $|\bar{D}| \ll |D|$. After exploring various subsampling strategies, we found uniform random sampling as effective as more complex methods (see \cref{app:add-exp:extract-ste}) and adopted it for simplicity. Since dataset reduction is not our main focus, we leave further optimizations for future work.

Reducing the dataset speeds up evaluation but does not guarantee identical results -- particularly when the subset is significantly smaller, as in our case. To bridge this gap, we build an IRT-based estimator that adjusts for this discrepancy, effectively estimating performance to reflect full-dataset results \cite{lord1968statistical, tinybenchmarks}.

\label{sec:estimate}

\paragraph{IRT model.}\label{sec:irt}
We first define an {estimator} to assess each endpoint model's inherent abilities, derived from the latents of a Bayesian network. This ensures that merging preserves individual model strengths. In the Evolve step (\S \ref{sec:evolve}), the estimated latent abilities are fed to a {\em performance} estimator to compute the final fitness.

To estimate LLM abilities, we build on \citet{tinybenchmarks}, who applied IRT to evaluate LLM performance; however, while they used IRT for benchmarking, we extend it to estimate inherent abilities relevant for model merging, and explicitly use them to guide merging in the Evolve step.

In IRT, latent variables ($\gamma$) represent a model's underlying abilities, while manifest variables ($Y$) indicate response correctness. The framework models the probability of a correct response based on model abilities and item characteristics (e.g., difficulty).

IRT defines this probability as:
\begin{equation}
     \mathbb{P}(Y_{im} = 1 \, | \, \gamma_m, \alpha_i, \beta_i) = \frac{1}{1 + \exp(-\alpha_i^\top \gamma_m + \beta_i)} \,
    \label{eq:irt}
\end{equation}
Here, ${\gamma_m \in \mathbb{R}^d}$ represents model $m$'s latent abilities, $\alpha_i \in \mathbb{R}^d$ defines the ability dimensions needed to answer example $i$, and $\beta_i$ denotes its difficulty. 
A model is more likely to answer correctly when its abilities ($\gamma_m$) align with the example’s required traits ($\alpha_i$) and less likely when the difficulty ($\beta_i$) is higher.
$Y_{im}$ is a binary variable indicating whether model $m$ correctly predicts example $i$ (1 if correct, 0 otherwise).

Crucially, this approach estimates a model’s likelihood of answering correctly {\em without directly analyzing the example's content}, relying solely on the estimated IRT parameters ($\gamma_m, \alpha_i, \beta_i$).

\paragraph{Fitting.}\label{sec:fit_irt}
We use variational inference to efficiently estimate both example-specific $(\alpha_i, \beta_i)$ and model-specific ($\gamma_m$) parameters within a hierarchical Bayesian model \citep{lalor2023py}, initialized as detailed in appendix \ref{app:fitting_details}.
Following \citet{tinybenchmarks}, we estimate $\alpha_i$ and $\beta_i$ using correctness data ($Y_{im}$) from publicly available model evaluations, namely the Open LLM leaderboard. 
To estimate $\gamma_m$, each endpoint model generates answers for the full evaluation dataset, which are then used to assess correctness ($Y_i$) (see Figure~\ref{fig:teaser}). This procedure is repeated for each model $m$, producing the corresponding $\gamma_m$ ($\gamma_1$ and $\gamma_2$ in the Figure).

To summarize, unlike previous work, where IRT latent abilities remain hidden variables, we explicitly derive $\gamma_m$ as an {\em ability estimator} to quantify each model's strengths. Additionally, rather than estimating $\gamma_m$ from a subset, we compute it using the {\em full} evaluation dataset, providing a more comprehensive measure of model ability, which we now leverage to enhance the merging process.

\begin{figure*}
    \centering
    \includegraphics[width=\textwidth]{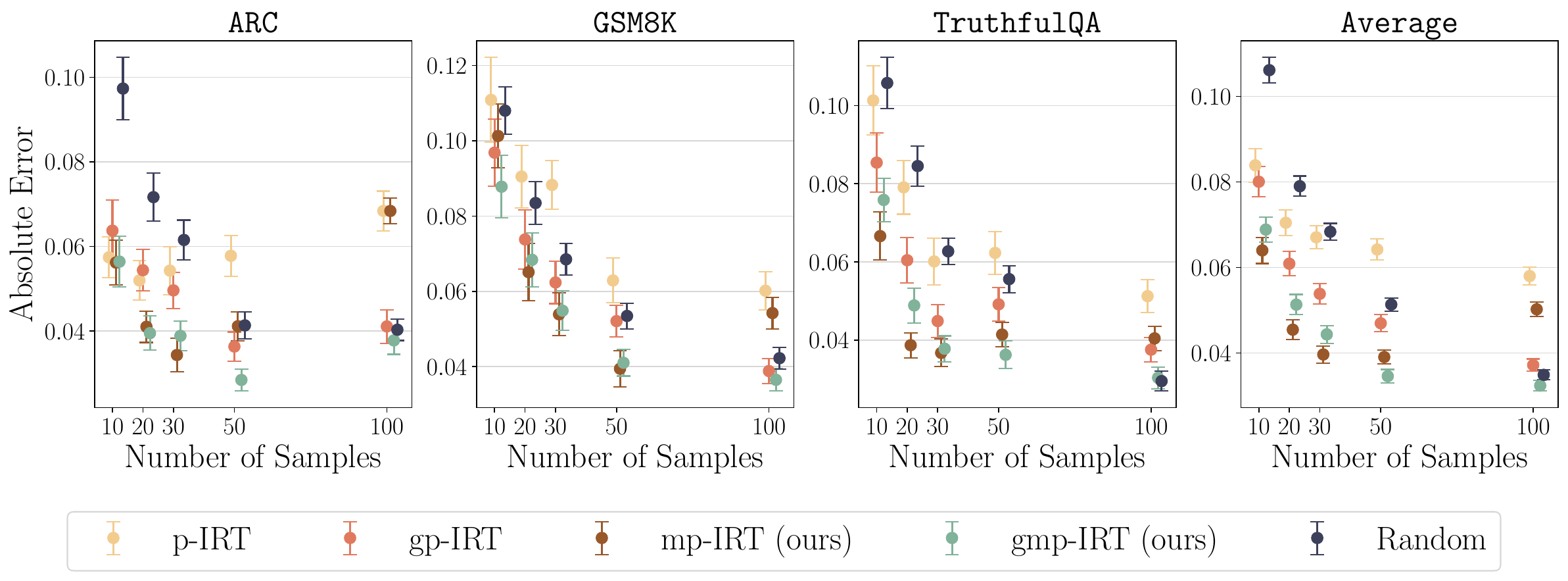}
    \caption{\textit{Performance Estimators:} Absolute error of various estimators as a function of sample size (lower is better). Our \mpirt{} and \gmpirt{} estimators consistently achieve lower error across various sample sizes and datasets.
     Additional results available in \Cref{fig:estimation_comparison_winogrande_hellaswag}.}
    \label{fig:estimation_comparison}
\end{figure*}

\subsection{Evolve: Performance Estimator}
\label{sec:evolve}
The \textit{performance estimator}, a key part of the Evolve step, efficiently approximates the fitness function, which measures the merged model's accuracy. Since fitness evaluation runs repeatedly during evolution (once per model per iteration), reducing its computational cost is crucial. Instead of evaluating the full dataset, the estimator predicts performance using only the endpoint models' abilities and the reduced dataset from previous steps, significantly accelerating the process.

We introduce two novel performance estimators for merging: \mpirt{} and \gmpirt{}. Since model merging linearly combines weights, we assume the merged model's ability is also a linear combination of the endpoint abilities. This makes our approach far more efficient, estimating only the interpolation coefficients ($\lambda_i$) instead of recomputing the full ability vector $\gamma$ of the merged model from scratch (as done in \pirt{} and \gpirt{} \cite{tinybenchmarks}).
\begin{assumption}[Linear Combination of Latent Abilities]\label{conj:latent-abilities-linear}
Let \(\{m_0, m_1, \dots, m_n\}\) be endpoint models with latent ability vectors \(\gamma_{i}\). If a new model \(\tilde{m}\) is formed as a linear combination of their parameters, its ability vector \(\gamma_{\tilde{m}}\) can be expressed as:
\begin{align}
    \gamma_{\tilde{m}}
    \;=\;
    \sum_{i=1}^n \lambda_i \,\gamma_{i}
    \;=\;
    [\gamma_{1}, \ldots, \gamma_{n}]\,\lambda
\end{align}
where \(\lambda = (\lambda_1, \ldots, \lambda_n)\) are the interpolation coefficients.
\end{assumption}
This assumption allows us to compute the multidimensional IRT model (Eq. \ref{eq:irt}) for model merging as a linear combination of the individual models' abilities:
\begin{align}
    p_{i\tilde{m}} 
    &= \mathbb{P}\!\left( Y_{i\tilde{m}} = 1 \;\middle|\; 
    \lambda_1 \gamma_{1} + \lambda_2 \gamma_{2}, \, 
    \alpha_i, \, \beta_i \right) \nonumber\\
    &= \frac{1}{1 + \exp\!\left( 
    -\alpha_i^\top \bigl( \lambda_1 \gamma_{1} 
    + \lambda_2 \gamma_{2} \bigr) 
    + \beta_i \right)}
     \label{eq:prob_model}
\end{align}
Since the endpoint models' latent abilities \(\gamma_i\) were pre-estimated over the full dataset \(D\) in the Estimate step, we only need the subset \(\bar{D}\) to estimate the interpolation coefficients \(\lambda_i\) via MLE.

\paragraph{Performance Estimators.}
To estimate the accuracy of the merged model $\tilde{m}$ using only the reduced dataset $\bar{D}$ and $p_{i\tilde{m}}$, we define the \emph{merged performance-IRT} (\textsc{mp-IRT}) estimator as:
\begin{align}
    \hat{Z}_{\tilde{m}}^{\mathrm{mp\text{-}IRT}}=
    \frac{\hat{\tau}}{\lvert \bar{D} \rvert} 
    \sum_{i \in \bar{D}} 
    Y_{i\tilde{m}}
    +
    \frac{1 - \hat{\tau}}{\lvert D \setminus \bar{D} \rvert} 
    \sum_{i \in D \setminus \bar{D}} 
    \hat{p}_{i\tilde{m}}
\end{align}
where ${\hat{\tau} = \frac{|\bar{D}|}{|D|}}$ downweights smaller subsets that may be noisier. In practice, we are considering the observed correctness for the data points we have access to, while $\hat{p}_{i\tilde{m}}$ predictions are used for the rest, enabling accurate performance estimation across all examples despite evaluating only a subset, where $\hat{p}_{i\tilde{m}} = \mathbb{P}\!\left( Y_{i\tilde{m}} = 1 \;\middle|\; \hat{\lambda}_1 \hat{\gamma}_{1} + \hat{\lambda}_2 \hat{\gamma}_{2}, \, \hat{\alpha}_i, \, \hat{\beta}_i \right)$ is the distribution defined by plugging into \cref{eq:prob_model} the parameter found via MLE. 

Although designed for model merging, \(\hat{Z}_{\tilde{m}}^{\mathrm{mp\text{-}IRT}}\) inherits certain limitations of \(\textsc{p-IRT}\)~\citep{tinybenchmarks}, such as non-uniform weighting and imperfect IRT fits. To mitigate these, we define a \emph{generalized} estimator that interpolates between \(\hat{Z}_{\tilde{m}}^{\mathrm{mp\text{-}IRT}}\) and the observed correctness on \(\bar{D}\):
\begin{align}\label{gp-irt}
    \hat{Z}_{\tilde{m}}^{\mathrm{gmp\text{-}IRT}}
    \;=\;
    c \sum_{i \,\in\, \bar{D}} w_i\,\hat{Y}_{i\tilde{m}}
    \;+\;
    (1 - c)\;\hat{Z}_{\tilde{m}}^{\mathrm{mp\text{-}IRT}} 
\end{align}
where \(c\) is a heuristic scalar chosen as in~\citet{tinybenchmarks} and \(w_i\) are uniform per-sample weights.

Although model merging can sometimes degrade performance due to weight interference—suggesting non-linear ability interactions— our assumption is empirically supported as we are interested only in evolved models that show a positive performance gain. As validated in our experiments (\S \ref{sec:exp_validation}), our custom estimators, designed around this assumption, outperform standard IRT estimators.

\subsection{Evolve: Evolutionary Search}
\label{sec:evolve_evo}  
The final step of our algorithm frames model merging as a multi-objective optimization problem.  
Each merging objective \( F(\tilde{m}, D_i) \) represents the performance of the merged model \(\tilde{m}\) on task~\(i\). In practice, we select a multi-objective evolutionary algorithm (e.g., NSGA-II~\citep{nsga-ii}) and a merging strategy (e.g., TIES~\citep{ties}), aiming to optimize the corresponding {Pareto front}, formally defined as:  
\[
P_{\overline{F}_D}(\Theta)
\;=\;
\bigl\{
\theta_i \,\in\, \Theta 
:\; \nexists\,\theta_j \,\in\, \Theta
\;\text{s.t.}\;\theta_j \succ \theta_i
\bigr\}
\]  
where \(\succ\) denotes \emph{Pareto-dominance}. A model \(m\) Pareto-dominates \(m'\) if:  
\begin{align*}
\forall \, F \,\in\, \overline{F}_D:\; 
F(m; D) &\leq F(m'; D)\\
&\quad\text{and}\\
\exists \, F \,\in\, \overline{F}_D:\; 
F(m; D) &< F(m'; D)
\end{align*}  
This means \(m\) is strictly better in at least one metric and no worse in all others. Models on the Pareto front are thus not dominated by any other model.

In our setting, to reduce computational costs, we approximate optimization using \(\overline{F}_{\bar{D}}\) instead of \(\overline{F}_D\), where \(\bar{D} \subset D\) is obtained by the \textit{extraction} step. Performance on \(\bar{D}\) is then estimated using the {performance estimator}.

%% file: 4_Experiments/content.tex
\begin{figure}
    \includegraphics[width=\linewidth]{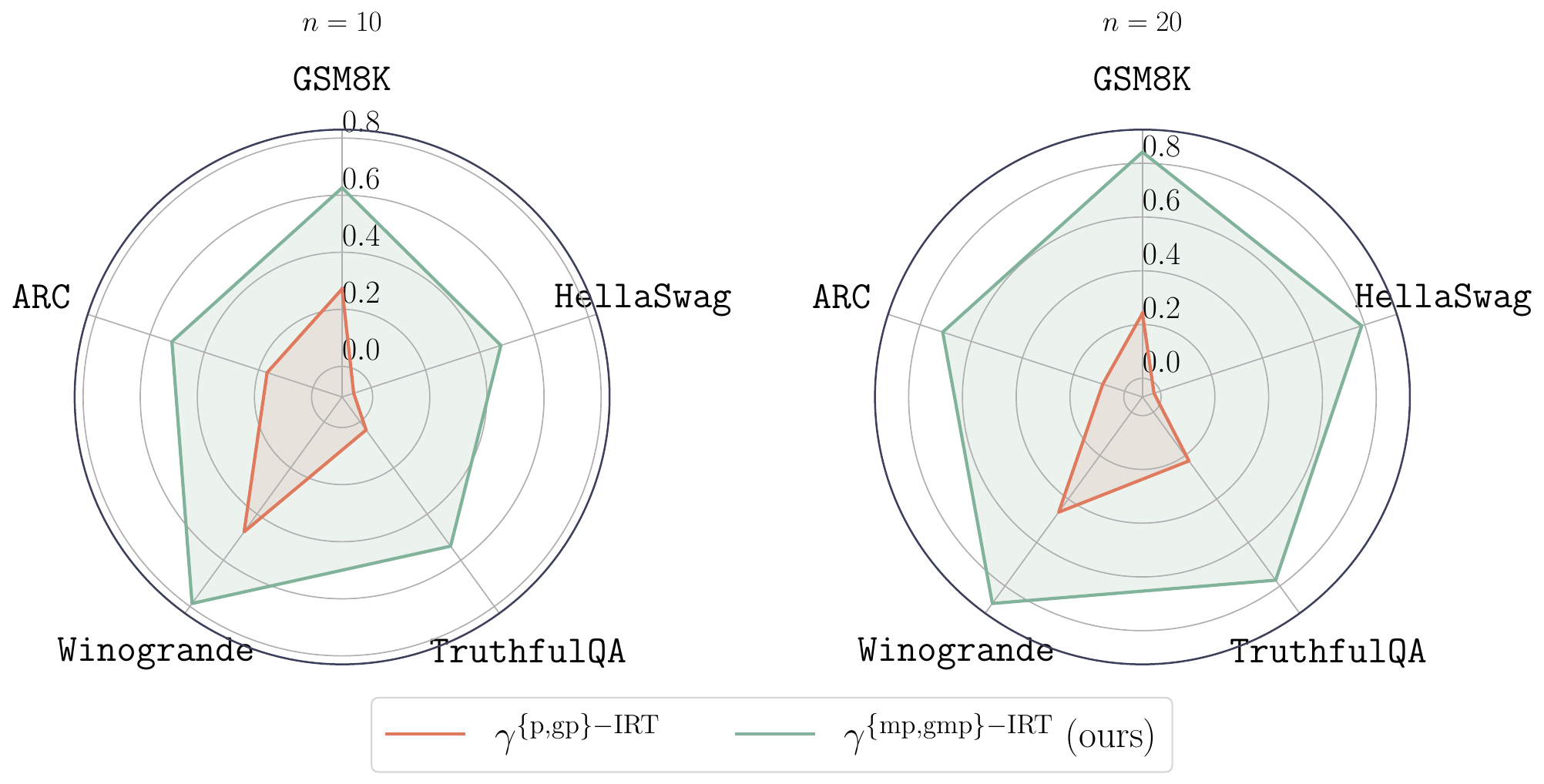}
    \caption{\textit{Ability Estimator:} Cosine similarity between estimated and true abilities for different tasks (higher is better). Our estimated abilities $\gamma^{\{\mathrm{mp},\mathrm{gmp}\}-{\mathrm{IRT}}}$ better approximate true abilities.}
    \label{fig:ability-estimator-comparison}
\end{figure}

\begin{figure*}
    \centering
    \begin{subfigure}{0.33\linewidth}
    \includegraphics[width=\linewidth]{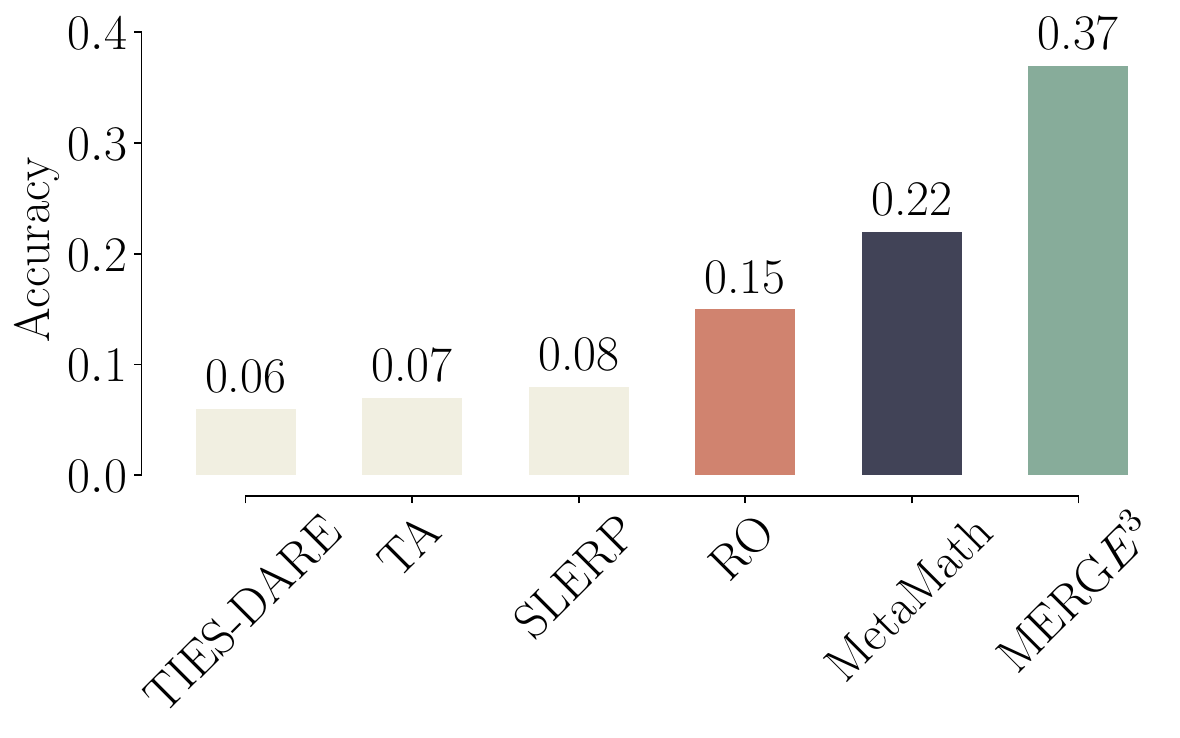}
    \caption{To Romanian.}
    \label{fig:cross-ro}
    \end{subfigure}
    \hfill
    \begin{subfigure}{0.33\linewidth}
    \includegraphics[width=\linewidth]{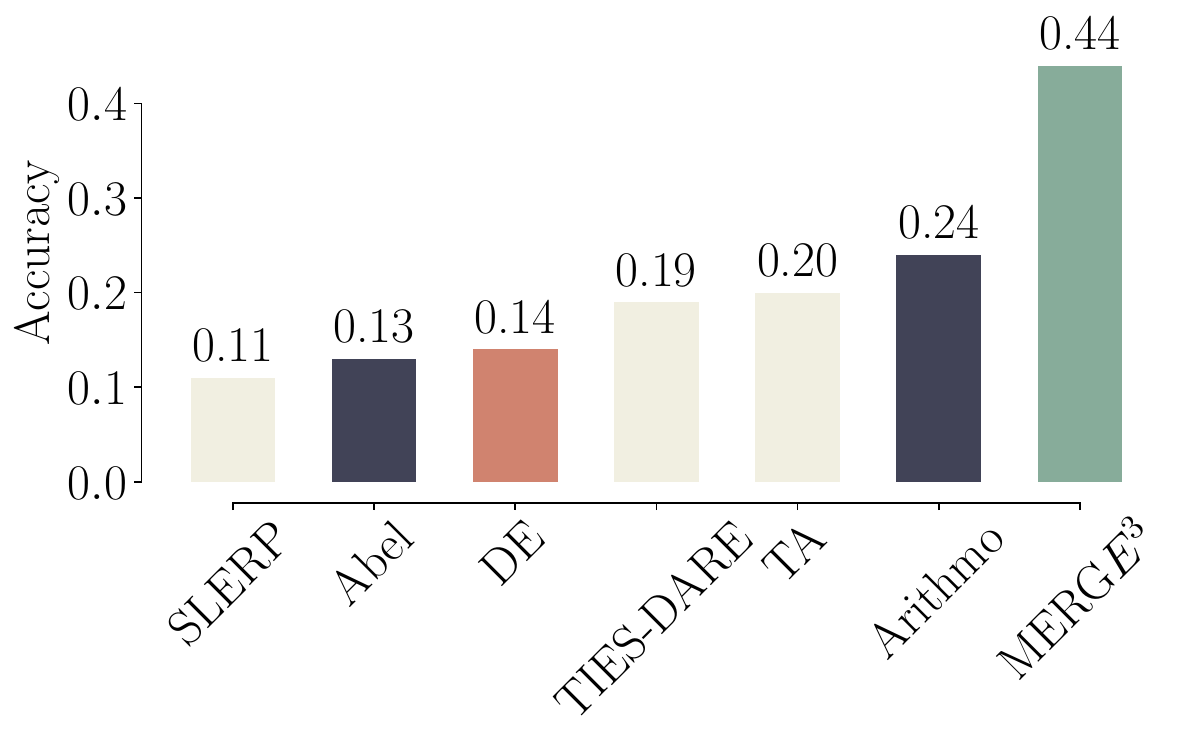}
    \caption{To German.}
    \label{fig:cross-de}
    \end{subfigure}
    \begin{subfigure}{0.33\linewidth}
    \includegraphics[width=\linewidth]{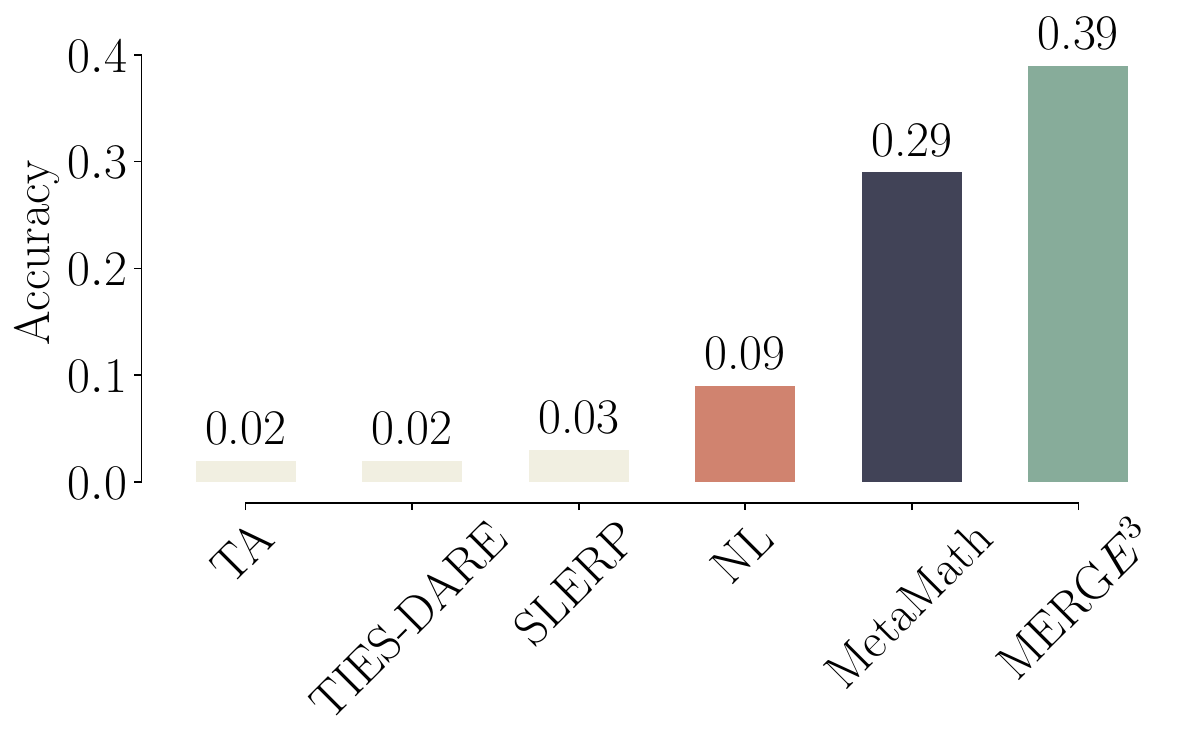}
    \caption{To Dutch.}
    \label{fig:cross-nl}
    \end{subfigure}
    \caption{\textit{Cross-lingual skill transfer}: merging math models (dark blue) with language-specific models (red) effectively transfers mathematical skills across languages (green - our method) compared to baselines (white). Accuracy on \dataset{GSM8K} for each target language.
    %, with white indicating baseline merging approaches.
    }
    \label{fig:cross-transfer}
\end{figure*}

\begin{figure}
    \centering
    \includegraphics[width=.9\linewidth]{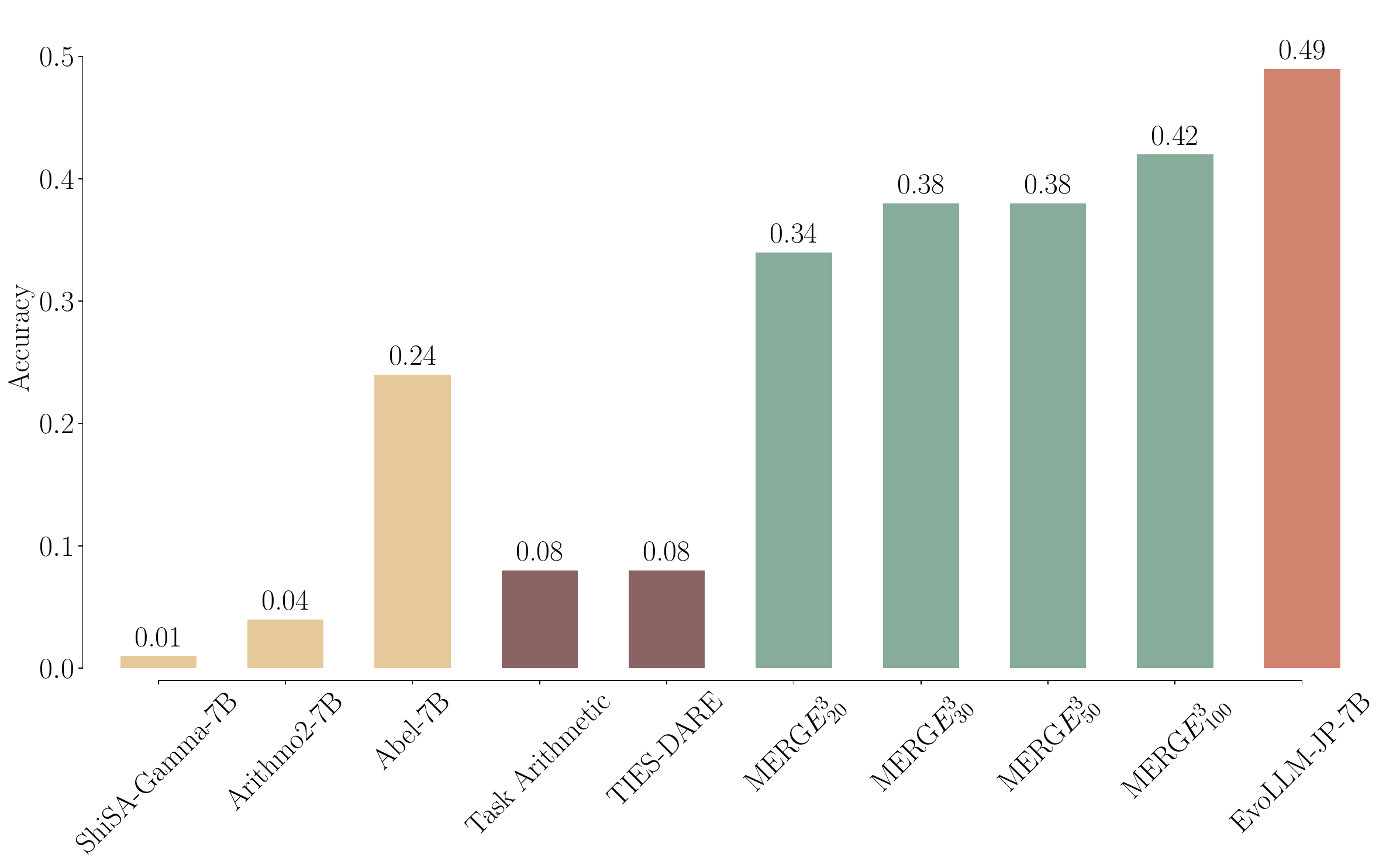}
    \caption{Accuracy of merged models for Japanese \dataset{GSM8K}.}
    \label{fig:evolve-comparison-sakana}
\end{figure}
In this section, we evaluate \approachns, demonstrating its effectiveness in evolutionary model merging on consumer-grade GPUs. We first validate the proposed ability and performance estimators, assessing their accuracy in approximating full-dataset evaluations. Next, we examine cross-lingual transfer, where \approach enables efficient merging of multilingual models, improving mathematical reasoning across languages. Finally, we evaluate its ability to synthesize multilingual models, surpassing individual fine-tuned baselines while remaining computationally efficient.
All the merging experiments were performed with our custom-made library \textit{Mergenetic} (see Appendix \ref{app:mergenetic}).

\subsection{Validating Estimators}
In this section, we empirically validate our merged-performance estimators by comparing them against standard \pirt{} and \gpirt{} estimators \cite{tinybenchmarks} across five benchmark datasets: \dataset{GSM8K} \cite{gsm8k}, \dataset{Winogrande} \cite{sakaguchi2021winogrande}, \dataset{TruthfulQA} \cite{lin2021truthfulqa}, \dataset{Hellaswag} \cite{zellers2019hellaswag}, and \dataset{ARC} \cite{ARC}. Due to space limitations, additional results are provided in Appendix \ref{app:add-exp}.

\paragraph{Ability Estimators.} 
\label{sec:exp_validation}
To validate our ability estimators we compare their inferred latent ability vectors to the reference ``ground-truth'' vectors \(\Gamma\). Specifically, we measure the cosine similarity and the Euclidean distance from the ground-truth \(\Gamma\) both for \(\gamma^{{\{\mathrm{mp,gmp}\}-\mathrm{IRT}}}\), estimated with our merged-performance IRT approaches, and \(\gamma^{\{\mathrm{p,gp}\}-\mathrm{IRT}}\), estimated with the \pirt{} and \gpirt{} estimators \citep{tinybenchmarks}.
Here, \(\Gamma_{m}\) is computed by fitting the IRT model (as in \cref{sec:irt}) to each merged model \(m\) using its entire set of responses on the full dataset \(D\). Incorporating all available data, \(\Gamma_{m}\) serves as our best proxy for the model’s true ability. Conversely, both \(\gamma_m^{{\{\mathrm{mp,gmp}\}-\mathrm{IRT}}}\) and \(\gamma_m^{\{\mathrm{p,gp}\}-\mathrm{IRT}}\) are estimated using only a smaller subset \(\bar{D} \subset D\) of size \(n\).
\Cref{fig:ability-estimator-comparison} shows the results of this comparison for $n=10$ and $n=20$, while the results for $n=15, 30, 50, 100$ are reported in \Cref{app:estimation-step} along with the same experiment over different languages. 
Across all five benchmark tasks our proposed ability estimator \(\gamma_m^{{\{\mathrm{mp,gmp}\}-\mathrm{IRT}}}\) consistently yields ability vectors with higher cosine similarity to \(\Gamma\) than \(\gamma_m^{\{\mathrm{p,gp}\}-\mathrm{IRT}}\). This trend is evident across both subset sizes, highlighting the robustness of our approach even with limited data. The superior performance of \(\gamma_m^{{\{\mathrm{mp,gmp}\}-\mathrm{IRT}}}\) empirically validates \Cref{conj:latent-abilities-linear}, confirming that an IRT-based ability estimator designed around this assumption provides more accurate ability estimates than a general-purpose alternative.

\paragraph{Performance Estimators.} \label{par: pe}
To assess the accuracy of our proposed performance estimators, we measure their absolute estimation error across different sample sizes. Specifically, we evaluate the performance estimates of six merged models using random sampling, \pirt{}, \gpirt{} \cite{tinybenchmarks}, \mpirt{}, and \gmpirt{} across various subset sizes. The resulting absolute errors shown in Figure~\ref{fig:estimation_comparison} are reported for \dataset{ARC}, \dataset{GSM8K}, \dataset{TruthfulQA}, and an aggregate average across all five benchmarks.

As shown in the figure, our proposed estimators, \mpirt{} and \gmpirt{}, consistently achieve lower absolute error compared to \gpirt{} and \pirt{}. While all IRT-based methods outperform random sampling, the incorporation of merged-performance IRT significantly enhances estimation accuracy. Notably, both \mpirt{} and \gmpirt{} maintain low empirical error and reduced variance even when operating with very small subsets (\(|\bar{D}| \approx 1.5\%\) of the full dataset). This highlights the robustness of our approach in low-data regimes. 

Since lower empirical error often correlates with reduced \emph{expected} error (as formalized in Section~\ref{sec:theory}), we adopt \mpirt{} and \gmpirt{} as our primary estimators for evolving merged language models in subsequent experiments.

% Math transfer to DE, RO, NL, Jap
\subsection{Cross-Lingual Transfer of Mathematical Skills}
\label{sec: cross-lingual}
To assess the transfer of mathematical reasoning from English to other languages, we merge an English math-specialized model with a \model{Mistral-7B} \cite{mistral} fine-tuned on each target language, then evaluate on the corresponding \dataset{GSM8K} translations \citep{gsm8k}. Appendix \ref{app:add-details-evo} provides details on the specific models used for merging. Following \citet{sakana}, we label an answer correct only if it is both accurate and written in the target language. We benchmark our approach against three commonly used merging baselines -- \method{Task Arithmetic} \citep{task-vectors}, \method{TIES} \citep{ties} and \method{DARE} \citep{yu2024language}. Following standard practice in the merging community, we apply either \method{TIES} and \method{DARE} jointly or \method{SLERP} \cite{10.1145/325165.325242}.

As shown in figure \ref{fig:cross-transfer}, merging a language-specific fine-tuning with a math-specialized model consistently surpasses both endpoint models by 10–20\% in accuracy on the translated \dataset{GSM8K}. In contrast, standard baselines often yield sub-optimal merges, performing worse than the endpoints themselves. This highlights the importance of optimized merging coefficients and motivates our evolutionary framework. To rule out the possibility that gains arise merely from exploiting a small, in-distribution dataset—rather than true cross-lingual transfer—we also merge the same Italian fine-tuning with itself; the results, available in the appendix \ref{app:add-exp-self-merging}, shows no improvement, reinforcing that our observed gains stem from genuine knowledge transfer.

Next, we evaluate our method for transferring math skills from English to Japanese and compare it to \method{EvoMerge} \citep{sakana}, which serves as an upper bound by computing fitness on the full dataset. As illustrated in figure \ref{fig:evolve-comparison-sakana}, our approach confirms the significant gains seen for the other languages, greatly surpassing both the performance of the endpoint models and that of the merging baselines.
While the accuracy is lower than that of the model obtained by computing the fitness on the full dataset as done by \citet{sakana}, figure \ref{fig:speed_accuracy} shows that our approximation yields a method that is $50\times$ more efficient, effectively making evolutionary merging feasible on a single consumer GPU.

\subsection{Evolving a Multilingual model}
\label{sec: multi-lingual}

\begin{table}
      \caption{Evolving a multilingual model. For each language, we report the accuracy on the corresponding translated \dataset{ARC} of both the language-specific model and the evolved multilingual model.} 
      \label{tab:multilingual}
  \resizebox{\linewidth}{!}{%
    \begin{tabular}{ccccccccc}
    \\
    \toprule
    \multirow{2}{*}{Model} & \multicolumn{8}{c}{\textbf{Accuracy} ($\uparrow$)} \\ 
    \cmidrule{2-9}
    & \multicolumn{2}{c}{Italian} & \multicolumn{2}{c}{English} & \multicolumn{2}{c}{German} & \multicolumn{2}{c}{Dutch} \\
    \midrule
    Finetuned & 0.61 & -- & 0.75 & -- & 0.61 & -- & 0.50 & -- \\ 
    \approach & \textbf{0.69} & \improvUP{8\%} & \textbf{0.79} & \improvUP{4\%} & \textbf{0.72} & \improvUP{11\%} & \textbf{0.69} & \improvUP{19\%} \\ 
    \bottomrule
    \end{tabular}
    }
\end{table}
We next combine individually fine-tuned models for \{\texttt{IT, EN, DE, NL}\} into a single multilingual model. Appendix \ref{app:add-details-evo} provides details on the specific models used for each language. As shown in \cref{tab:multilingual}, the resulting merged model surpasses each language-specific variant by up to 19\% in accuracy on the \dataset{ARC-Challenge} dataset \cite{ARC}. Even more notably, it outperforms all its constituent endpoints, demonstrating a clear positive transfer of knowledge across languages.
Beyond the clear accuracy boosts in each language, a few key insights stand out. First, the largest improvement occurs for Dutch (from 50\% to 69\%), suggesting that merging particularly benefits languages where the baseline performance is lower. Second, even English, which starts from the highest baseline, still gains by 4\%, indicating that positive transfer is not limited to low-resource or weaker endpoints. Finally, the fact that the merged model outperforms all individual fine-tunings (rather than landing between them) points to a genuine cross-lingual synergy, wherein knowledge from each language-specific model collectively strengthens the multilingual result.

%% file: 5_Theoretical_analysis/content.tex
\label{sec:theory}

In this section, we provide theoretical guarantees for our \textit{performance estimator}, demonstrating that its estimated accuracy is a reliable approximation of full-dataset accuracy. We provide formal guarantees for its performance, analyze its stability under dataset reduction, and explain why it remains a robust proxy for the true fitness of the merged models. 
This analysis not only solidifies the estimator's theoretical foundation but also offers practical insights into its behavior in finite-data and asymptotic regimes.

The section is structured as follows: first (\S \ref{subsec:eps-stability}), we derive a correlation between the accuracy of the performance estimator and the quality of the minimum found by solving an optimization problem using that performance estimator as objective function; second (\S \ref{subsec:mp-irt-theory}), we study the asymptotic properties of the performance estimator as the dataset size approaches infinity, formalizing it as an unbiased estimator; and finally (\S \ref{subsec:mp-irt-eps-stability-analysis}), we demonstrate that our performance estimator behaves in expectation within a \(\,\epsilon\)-bound of the accuracy on the true optimum dataset. The proofs for all the theorems and propositions presented below are outlined in \cref{app:proofs}.

\subsection{Part~I: \texorpdfstring{\(\epsilon\)}{ε}-Stable Estimators and 
\texorpdfstring{\(\epsilon\)}{ε}-Optimality Preservation}
\label{subsec:eps-stability}

We first consider a performance metric \(F(\theta;D)\) for 
\(\theta \in \Theta \subset \mathbb{R}^n\), where \(D\) is a dataset.  
If we choose a smaller subset \(\bar{D} \subset D\) to approximate this metric, 
denoted \(F(\theta;\bar{D})\), we wish to control the loss in optimality incurred 
by replacing \(F(\theta;D)\) with \(F(\theta;\bar{D})\).  

\begin{definition}[\(\epsilon\)-Stability.]
    Given two datasets \(D\) and \(\bar{D}\), we say \(F(\cdot;\bar{D})\) is 
    \emph{\(\epsilon\)-stable with respect to} \(F(\cdot;D)\) if, for all 
    \(\theta \in \Theta\),
    \[
      \bigl|F(\theta;D)\;-\;F(\theta;\bar{D})\bigr|\;\le\;\epsilon
    \]
\end{definition}

Under this condition, minimizing \(F(\cdot;\bar{D})\) yields an objective value 
within \(\epsilon\) of minimizing \(F(\cdot;D)\).  Formally:

\begin{theorem}[\(\epsilon\)-Optimality Preservation]
\label{thm:eps-opt-preserve}
Let \(D\) be a dataset, let \(\bar{D}\subset D\) be a subset, and let 
\(F(\cdot;\bar{D})\) be \(\epsilon\)-stable with respect to \(F(\cdot;D)\), 
with a fixed \(\epsilon>0\).  Define
\[
  \theta^\star
  \;=\;
  \arg\!\min_{\theta\in \Theta}\;F(\theta;D)
  \quad\text{and}\quad
  \hat{\theta}
  \;=\;
  \arg\!\min_{\theta\in \Theta}\;F(\theta;\bar{D})
\]
Then
\[
    \bigl|F(\theta^\star;D)\;-\;F(\hat{\theta};\bar{D})\bigr|
    \;\le\;
    \epsilon
\]
\end{theorem}
Thus, \(\epsilon\)-stability ensures that any global minimizer on \(\bar{D}\) 
achieves an objective value on \(D\) no worse than \(\epsilon\) from 
the true global optimum.
%
% \paragraph{\texorpdfstring{\(\epsilon\)}{ε}-Stability in Expectation.}
Nevertheless, uniformly bounding \(\bigl|F(\theta;D)-F(\theta;\bar{D})\bigr|\) for all \(\theta\) may be too strong in practice. For this reason, we introduce:
\begin{definition}[\(\epsilon\)-Stability in expectation]\label{subsec:eps-stability-expectation}
    Given two datasets \(D\) and \(\bar{D}\), we say \(F(\cdot;\bar{D})\) is 
    \emph{\(\epsilon\)-stable in expectation with respect to} \(F(\cdot;D)\) if
    \[    \mathbb{E}_{\bar{D}}\bigl[\bigl|F(\theta;D)\;-\;F(\theta;\bar{D}) \bigr|\bigr]
      \;\le\;\epsilon
    \]
    where the expectation is over the (random) choice of \(\bar{D}\)
\end{definition}
Under this  relaxed notion, we still obtain a similar control on the \emph{expected} suboptimality gap:
\begin{theorem}[Expected \(\epsilon\)-Stability of the Minimum]
\label{thm:expected-eps-stability}
Suppose \(F(\cdot;\bar{D})\) is \(\epsilon\)-stable in expectation with respect 
to \(F(\cdot;D)\).  Let 
\[
  m^\star \;:=\;\min_{\theta\in\Theta}\,F(\theta;D)
  \quad\text{and}\quad
  \widehat{m}(\bar{D}) \;:=\;\min_{\theta\in\Theta}\,F(\theta;\bar{D})
\]
Then
\[
  \bigl|
    m^\star
    \;-\;
    \mathbb{E}_{\bar{D}}\bigl[\widehat{m}(\bar{D})\bigr]
  \bigr|
  \;\le\;
  \epsilon
\]
\end{theorem}

Hence, even if stability only holds \emph{on average}, the expected gap between 
the global optimum on \(D\) and the optimum on \(\bar{D}\) remains at most \(\epsilon\).

\subsection{Part~II: Theoretical Guarantees for \texorpdfstring{\mpirt{}}{mp-IRT}}
\label{subsec:mp-irt-theory}

We now apply these ideas to our proposed \mpirt{} estimator  (cf.\ \S\ref{sec:estimate}).  We first show that \mpirt{} is asymptotically  unbiased, and then combine this fact with \Cref{thm:expected-eps-stability}  to argue that \mpirt{}-based minimizers remain close to those that minimize  the full-dataset performance measure.

\paragraph{Asymptotic unbiasedness.}
\label{par:asymptotic-consistency}
The following proposition establishes that, as \(\bar{D}\) grows,  \(\hat{Z}^{\mathrm{mp\text{-}IRT}}\) converges in probability to the true performance \(Z\). Its proof relies on classical limit arguments for unbiased estimators.

\begin{proposition}[Asymptotic unbiasedness of \mpirt{}]
\label{prop:estimator-unbiased}
Assume:
\begin{enumerate*}[label=(\roman*)]
    \item \(\hat{\lambda} \to \lambda\) in probability as 
    \(\lvert \hat{I}\rvert \to \infty\),
    \item for each \(i\in I\), the true values \(\alpha_i,\beta_i,\theta_1,\theta_2\) 
          are known, with \(\sup_{i\in I}\|\alpha_i\|_2 \le c\) for a fixed \(c\),
    \item linear inheritance of abilities 
          (cf.\ \Cref{conj:latent-abilities-linear}) holds.
\end{enumerate*}
Then, for all \(j,l\),
\begin{align*}
    &\Bigl|\mathbb{E}\bigl[\hat{Z}_{jl}\,\bigm|\,
      Y_{i_0l},\dots,Y_{i_kl}
    \bigr]
    \;-\;
    \mathbb{E}\bigl[Z_{jl}\,\bigm|\,
      Y_{i_0l},\dots,Y_{i_kl}
    \bigr]
  \Bigr| \;\to\; 0
\end{align*}
\end{proposition}
in probability as $\lvert \hat{I}\rvert \to \infty$.
Thus, for sufficiently large subsets \(\bar{D}\), the discrepancy between 
\(\hat{Z}_{\tilde{m}}\) and \(Z_{\tilde{m}}\) can be made arbitrarily small 
with high probability.

\subsection{Part~III: performance preservation via \mpirt{}}
\label{subsec:mp-irt-eps-stability-analysis}
We now conclude that \mpirt{} preserves near-optimality when we train on 
a suitably large \(\bar{D}\subset D\).  Since 
Proposition~\ref{prop:estimator-unbiased} asserts that \(\hat{Z}\) approximates 
\(Z\) well for large \(\lvert \bar{D}\rvert\), it follows (under mild conditions) 
that \(\mpirt{}\) remains \(\epsilon\)-stable in expectation.  Hence, 
\Cref{thm:expected-eps-stability} shows that minimizing \(\hat{Z}\) on 
\(\bar{D}\) yields, on average, a solution within \(\epsilon\) of the 
full-dataset optimum.

\begin{theorem}[Asymptotic performance preservation of \mpirt{}]
\label{thm:mpirt-asymptotic-eps-preserve}
Let \(\bar{D}\subset D\) be a random subset used to compute 
\(\hat{Z}^{\mathrm{mp\text{-}IRT}}\).  Suppose that, as \(\lvert \bar{D}\rvert\to\infty\), 
\(\hat{Z}^{\mathrm{mp\text{-}IRT}}\) converges in probability to \(Z\) (the true 
performance on \(D\)), and that \(\hat{Z}^{\mathrm{mp\text{-}IRT}}\) is 
\(\epsilon\)-stable in expectation for sufficiently large \(\lvert \bar{D}\rvert\).  
Then the expected global optimum of \(\hat{Z}^{\mathrm{mp\text{-}IRT}}\) on \(\bar{D}\) 
differs from that of \(Z\) on \(D\) by at most~\(\epsilon\).  As 
\(\lvert \bar{D}\rvert\to \infty\), \(\epsilon\to 0\).
\end{theorem}

\paragraph{Finite-sample analysis via the Law of Large Numbers.}
In practice, we rarely have \(\lvert \bar{D}\rvert\to\infty\).  Instead, one can 
appeal to \emph{expected} \(\epsilon\)-stability 
(\Cref{thm:expected-eps-stability}) and then \emph{estimate} the corresponding 
expectation empirically.  For instance, one may draw multiple subsets 
\(\bar{D}_1,\ldots,\bar{D}_S\) at random from \(D\) and compute 
\[
  \frac{1}{S}\,\sum_{s=1}^S 
  \Bigl\lvert F(\theta;D) \;-\; F(\theta;\bar{D}_s)\Bigr\rvert
\]
as an empirical approximation to 
\(\mathbb{E}_{\bar{D}}\bigl[\lvert F(\theta;D)-F(\theta;\bar{D})\rvert\bigr]\).
By the Law of Large Numbers, if this empirical average remains small (say, 
\(\approx \tilde{\epsilon}\)), then the true expectation is also small.  
Consequently, \Cref{thm:expected-eps-stability} implies that the optimal 
solution on each \(\bar{D}_s\) is within \(\tilde{\epsilon}\) of the global 
optimum on \(D\), on average.

\paragraph{Conclusion.}
In summary, \mpirt{} inherits asymptotic consistency from \(\mathrm{p\text{-}IRT}\) 
while requiring only a subset \(\bar{D}\subset D\).  By showing it is 
\(\epsilon\)-stable (in expectation) for large \(\lvert \bar{D}\rvert\), we conclude 
that \emph{optimizing on \(\bar{D}\) yields (on average) a solution close to the 
true optimum on \(D\).}  In finite-sample regimes, multiple random draws of 
\(\bar{D}\) can be used to empirically verify that the discrepancy remains small, 
thereby justifying the practical use of \mpirt{} on moderately sized subsets.

%% file: 7_Conclusions/content.tex
We introduced \approachns{}, an evolutionary merging framework that makes high-quality model merging feasible on a single consumer GPU. By combining a subset-based approach with IRT-driven performance estimation, \approachns{} reduces merging costs by up to fifty-fold compared to prior methods -- without sacrificing the quality of the merged model. Our experiments demonstrate successful cross-lingual transfer in mathematics (e.g., from English to Japanese), as well as the synthesis of new multilingual models that outperform each of their language-specific endpoints. Overall, \approachns{} expands the practical reach of evolutionary merging, allowing everyday practitioners to benefit from advanced multi-task and multilingual model compositions at a fraction of the usual computational cost.

\section*{Impact Statement}
The introduction of \approach provides an efficient and accessible method for evolutionary model merging on consumer-grade GPUs. By combining dataset reduction techniques and Item Response Theory (IRT)-based performance estimations, \approachns{} significantly lowers computational requirements while maintaining competitive performance. This enables researchers and developers to synthesize high-quality multilingual and cross-lingual models without requiring cluster-scale hardware.

The open-source release of \approach aims to make evolutionary model merging widely accessible, fostering further innovation in resource-constrained environments. With applications in multilingual NLP and low-resource language modeling, \approach addresses practical challenges in the field, offering an efficient solution for creating state-of-the-art models on standard hardware.

%% file: E_Library/content.tex
\label{app:mergenetic}
Each experiment was run using a library developed specifically for this paper, which will be released as open-source software, called \textit{Mergenetic}. This library allows for defining a merging problem as either a single-objective or multi-objective optimization problem, where one only needs to specify the merging method,a fitness function, and select the hyperparameters for a chosen evolutionary algorithm. 

The implementation relies on \textit{Mergekit }  \cite{goddard2025arceesmergekittoolkitmerging} for merging the models, \textit{Pymoo } \cite{Blank_2020} for optimizing the objective function through evolutionary algorithms, and  \textit{Lm-Evaluation-Harness
 } \cite{eval-harness} for implementing some of the fitness functions. In \cref{tab:merge_methods} we outline the supported merging methods, while in \cref{tab:algorithms} we outline the currently available evolutionary algorithms.

We believe this library is a significant contribution as it facilitates evolutionary model merging and aligns well with the paper's approach, which aims to reduce computational burden. It can be a valuable tool for the community and for users interested in cross-lingual transfer or creating multilingual models for target low-resource languages.

\begin{table}
    \centering
    \caption{Overview of supported merging methods in Mergenetic.}
    \vspace{10pt}
    \resizebox{\linewidth}{!}{%
    \begin{tabular}{lcc}
    \toprule
    \textbf{Method} & \textbf{Multi-Model} & \textbf{Uses Base Model} \\ 
    \midrule
    Linear (Model Soups) & Yes & No \\ 
    SLERP & No & Yes \\ 
    Task Arithmetic & Yes & Yes \\ 
    TIES & Yes & Yes \\ 
    DARE (TIES) & Yes & Yes \\ 
    DARE (Task Arithmetic) & Yes & Yes \\ 
    \bottomrule
    \end{tabular}
    }
    \label{tab:merge_methods}
\end{table}

\begin{table}
    \centering
    \caption{Overview of supported Pymoo's evolutionary algorithms in Mergenetic.}
    \vspace{10pt}
    \resizebox{\linewidth}{!}{%
    \begin{tabular}{lccc}
    \toprule
    \textbf{Algorithm} & \textbf{Class} & \textbf{Objective(s)} & \textbf{Constraints} \\ 
    \midrule
    Genetic Algorithm  & GA & single & x \\ 
    Differential Evolution & DE & single & x \\ 
    Biased Random Key GA & BRKGA & single & x \\ 
    Nelder Mead & NelderMead & single & x \\ 
    Pattern Search & PatternSearch & single & x \\ 
    CMAES & CMAES & single &  \\ 
    Evolutionary Strategy & ES & single &  \\ 
    SRES & SRES & single & x \\ 
    ISRES & ISRES & single & x \\ 
    NSGA-II & NSGA2 & multi & x \\ 
    R-NSGA-II & RNSGA2 & multi & x \\ 
    NSGA-III & NSGA3 & many & x \\ 
    U-NSGA-III & UNSGA3 & many & x \\ 
    R-NSGA-III & RNSGA3 & many & x \\ 
    MOEAD & MOEAD & many &  \\ 
    AGE-MOEA & AGEMOEA & many &  \\ 
    C-TAEA & CTAEA & many & x \\ 
    SMS-EMOA & SMS-EMOA & many & x \\ 
    RVEA & RVEA & many & x \\ 
    \bottomrule
    \end{tabular}
    }
    \label{tab:algorithms}
\end{table}

%% file: A_Details/content.tex
\label{app:add-details}
This section provides additional implementation and experimental details that were not included in the main paper.

\subsection{IRT Fitting Details} \label{app:fitting_details}
As previously stated, we used the implementation from \citet{tinybenchmarks} and adopted their configuration settings. Specifically, we used $\gamma_m \sim N(\mu_{\gamma}\ones_d, 1/u_{\gamma}I_d)$, $\alpha_{i} \sim N(\mu_{\alpha}\ones_d, 1/u_{\alpha}I_d)$, and $\beta_{i} \sim N(\mu_\beta, 1/u_\beta)$.
Following \citet{tinybenchmarks}, we also applied (hyper)priors to the prior parameters using the software for fitting hierarchical Bayesian models \citep{lalor2023py}: $\mu_{\gamma} \sim N(0, 10)$, $u_{\gamma} \sim \Gamma(1, 1)$, $\mu_{\alpha} \sim N(0, 10)$, $u_{\alpha} \sim \Gamma(1, 1)$, $\mu_\beta \sim N(0, 10)$, and $u_\beta \sim \Gamma(1, 1)$. 
For both the model and example-specific parameters $\gamma_m$, $\alpha_{i}$, and $\beta_{i}$, we take their point estimates as the means of their respective variational distributions. 
The $\gamma$ model dimensionality is set to $15$ following the parameter choice suggested by \citet{tinybenchmarks}.

\begin{table}[ht]
\centering
\caption{Mistral-based models with shortened column headers and names. Role can be either E, M or B, referring to endpoint, merge or base model respectively. Spec refers instead to specialization, with \texttt{mth}, \texttt{ger}, \texttt{ita}, \texttt{jpn}, \texttt{dut} and \texttt{gen} referring to Math, German, Italian, Japanese, Dutch and General respectively. We finally have the author and model ID as per the Huggingface.}
\label{tab:short-mistral-models}
\vspace{10pt}
\resizebox{\columnwidth}{!}{% 
\begin{tabular}{ccll}
\toprule
\textbf{Role} & \textbf{Spec} & \textbf{Author} & \textbf{Model} \\
\midrule
E & \texttt{mth} & \emph{upaya07 }      & \model{Arithmo2-Mistral-7B}     \\
E & \texttt{mth,jpn} & \emph{SakanaAI }      & \model{EvoLLM-JP-v1-7B}     \\
E & \texttt{mth} & \emph{GAIR}          & \model{Abel-7B-002}      \\
E & \texttt{mth} & \emph{meta-math}     & \model{MetaMath-Mistral-7B}     \\
B & \texttt{gen}  &\emph{ mistralai}     & \model{Mistral-7B-v0.1}      \\
E & \texttt{ger} &\emph{ jphme  }          & \model{em\_german\_mistral\_v01}   \\
E & \texttt{ger} & \emph{LeoLM  }       & \model{leo-mistral-hessianai-7b}       \\
E & \texttt{ita} & \emph{DeepMount00}   & \model{Mistral-Ita-7b}       \\
E & \texttt{jpn} & \emph{augmxnt  }     & \model{shisa-gamma-7b-v1}       \\
E & \texttt{dut} & \emph{BramVanroy}    & \model{GEITje-7B-ultra}     \\
E & \texttt{ro} & \emph{OpenLLM-Ro}    & \model{RoMistral-7b-Instruct}     \\
M & \texttt{gen} &\emph{ chlee10 }      & \model{T3Q-Merge-Mistral7B}    \\
E & \texttt{gen} &\emph{ liminerity}    & \model{M7-7b}        \\
E & \texttt{gen} & \emph{yam-peleg}     & \model{Experiment26-7B}     \\
M & \texttt{gen} & \emph{PracticeLLM}   & \model{SOLAR-tail-10.7B-Merge-v1.0}    \\
E & \texttt{gen} &\emph{ upstage}       & \model{SOLAR-10.7B-v1.0}      \\
E & \texttt{gen} & \emph{Yhyu13}        & \model{LMCocktail-10.7B-v1}   \\
M & \texttt{gen} & \emph{FuseAI}        & \model{FuseChat-7B-Slerp}  \\
M & \texttt{gen} & \emph{FuseAI}        & \model{FuseChat-7B-TA}   \\
E & \texttt{gen} & \emph{FuseAI}        & \model{OpenChat-3.5-7B-Mixtral}  \\
E & \texttt{gen} & \emph{FuseAI}        & \model{OpenChat-3.5-7B-Solar}  \\
M & \texttt{gen} & \emph{jan-hq}        & \model{supermario-slerp-v3}\\
E & \texttt{gen} & \emph{jan-hq}        & \model{supermario-slerp-v2}\\
E & \texttt{gen} & \emph{jan-hq}        & \model{supermario-v2}\\
M & \texttt{gen} & \emph{superlazycoder}& \model{NeuralPipe-7B-slerp
}   \\
E & \texttt{gen} & \emph{OpenPipe}      & \model{mistral-ft-optimized-1218}     \\
E & \texttt{gen} & \emph{mlabonne}      & \model{NeuralHermes-2.5-Mistral-7B} \\
\bottomrule
\end{tabular}%
}
\end{table}

\subsection{\approachbf{} Algorithm}
Below we outline the pseudo-code for the end-to-end \approach{} algorithm:
\begin{algorithm}[ht]
    \caption{MERGE3 Algorithm} \label{alg:merge3}
    \begin{algorithmic}[1]
    \REQUIRE Dataset $D$, models $\lbrace M_1, M_2, \ldots, M_n \rbrace$, iterations $T$
    \ENSURE Pareto-optimal merged models
    % \PROCEDURE{MERGE3}{$D, \lbrace M_1, \ldots, M_n \rbrace, T$}
        \STATE $\bar{D} \gets \Call{RandomSample}{D, k}$
            \Comment{Sample $k$ items from $D$}
        \STATE $\lbrace \gamma_1,\gamma_2,\ldots,\gamma_n \rbrace \gets \Call{EstimateAbilities}{\{M_1, \ldots, M_n\}, D}$
        \STATE $P \gets \text{GenerateInitialPopulation}{\{M_1, \ldots, M_n\}}$
            \Comment{Initialize population}
        \FOR{$t \gets 1$ to $T$}
            \FORALL{$M \in P$}
                \STATE $\lambda \gets \Call{FitLambda}{M,\,\lbrace \gamma_1,\dots,\gamma_n \rbrace,\,\bar{D}}$
                    \Comment{Fit interpolation weights}
                \STATE $\text{preds} \gets \Call{GetPredictions}{M,\,\bar{D},\,\lambda}$
                    \Comment{Compute predictions}
                \STATE $\text{corr} \gets \Call{GetCorrectness}{\text{preds},\,\bar{D}}$
                    \Comment{Evaluate correctness}
                \STATE $F(M) \gets \Call{EstimateFitness}{\text{corr},\,\lambda}$
                    \Comment{Estimate fitness score}
            \ENDFOR
            \STATE $P \gets \Call{SelectParents}{P, f}$ 
                \Comment{Select based on fitness}
            \STATE $P \gets \Call{ApplyMutation}{P}$
                \Comment{Apply mutation}
            \STATE $P \gets \Call{ApplyCrossover}{P}$
                \Comment{Generate offspring}
        \ENDFOR
        \STATE \textbf{return} $\Call{ParetoFront}{P}$
    % \ENDPROCEDURE
    \end{algorithmic}
\end{algorithm}

\subsection{Experimental Details}
\label{app:add-details-evo}

\paragraph{Models.}
One key assumption of model merging is that the endpoint models lie within the same basin \cite{task-vectors}. This means that merging arbitrary models is not feasible; rather, \textbf{all models involved must be fine-tuned versions of the same base model}. To satisfy this requirement, we selected several fine-tuned models from the Hugging Face Hub that originated from the same base model. Specifically, we focused on models fine-tuned starting from \model{Mistral-7B} \cite{mistral}, following common best practices in the community \cite{sakana}.
\Cref{tab:short-mistral-models} lists all the models used in our experiments, along with their corresponding names on the Hugging Face Hub. A total of $27$ models were considered for our experiments.

In the rest of the section, we provide further details for reproducing the experiments in \cref{sec: cross-lingual} and \cref{sec: multi-lingual} of the main paper.

\subsubsection{Cross-Lingual Transfer}

In the cross-lingual transfer evolutionary merging, we evolved four merged models with mathematical capabilities in different languages: Japanese, Romanian, German, and Dutch. In each of these experiments, we deployed an ad-hoc genetic algorithm for single-objective optimization. We employed the Simulated Binary Crossover~\cite{10.1145/1276958.1277190} operator to generate offspring solutions by combining parent solutions. To maintain diversity and explore the search space, we applied Polynomial Mutation~\cite{10.1145/1276958.1277190}, which introduces small perturbations to offspring solutions and enhances the algorithm's ability to escape local optima. This combination of SBX and PM effectively balances exploration and exploitation, facilitating efficient convergence toward optimal solutions.

Furthermore, guided by empirical tests, we decided to deploy \method{SLERP} to evolve solutions for the Romanian and Dutch problems, while we used a combination of \method{TIES} and \method{DARE} for the Japanese and the German ones. We deployed four different sizes of the fitness datasets for Japanese, namely 20, 30, 50, and 100, in order to obtain a more detailed analysis of the method for comparison with the work of~\cite{sakana}. On the other hand, we kept the fitness dataset size fixed to 20 for all other aforementioned experiments. The fitness dataset was extracted from the test set of \dataset{GSM8K}, and we used the remaining, non-overlapping samples as test set for evaluating the model. To get the language-specific versions of \dataset{GSM8K}, we used \model{Unbabel/TowerInstruct-7B-v0.2} \cite{alves2024tower} to translate the datasets. In each experiment, the population size was fixed to 25 and the number of iterations to 7.

To check the correctness of the solution, following \citet{sakana}, we used a regex to extract the last numerical value returned in the model's answer and compare it with the ground truth. The solution is also checked to be in the correct language with the language identifier from fastText~\cite{joulin2017bag}.

The mathematical models used in combination with \method{TIES} and \method{DARE} were \model{Abel-7B-002} and \model{Arithmo2-Mistral-7B}, whereas we used \model{MetaMath-Mistral-7B} in combination with \method{SLERP}. Moreover, we employed the following language-specialized models: \model{shisa-gamma-7b-v1}, \model{em\_german\_mistral\_v01}, \model{GEITje-7B-ultra}, and \model{RoMistral-7b-Instruct}. More information about these models can be found in \cref{tab:short-mistral-models}.

Lastly, we evaluated \model{EvoLLM-JP-v1-7B} \cite{sakana} under the same conditions as \approach{} to assess its accuracy, following the prompting structure outlined by \citet{sakana}.

\subsubsection{Multi-Lingual Transfer}

In this experiment, we tackle the ARC dataset in multiple languages (Italian, Dutch, German, and English)\footnote{We used the dataset on the Hugging Face Hub from openGPT-X/arcx} \cite{thellmann2024crosslingual} using a multi-objective evolutionary merging procedure based this time on NSGA-II \cite{nsga-ii}. We configure the population size to 25 and the number of evolutionary iterations to 7. We deployed a combination of \method{TIES} and \method{DARE} as merging strategy. As in previous settings, both the fitness function and the test metrics operate by extracting the final model-generated choice via a regex, but this time they look for an instance from the set \{A, B, C, D\} rather than a number. On top of this, we employed a dataset composed by 20 datapoints for each language from the relative translation of \dataset{ARC} to compute the fitness, and we extracted the test set as for the previous experiments. Furthermore, unlike the single-objective approach described earlier, here we explicitly optimize multiple objectives simultaneously. This time, the employed models are \model{Mistral-Ita-7b}, \model{GEITje-7B-ultra}, \model{leo-mistral-hessianai-7b}, and the base model \model{Mistral-7B-v0.1}. 

\subsubsection{Ability and Performance Estimator} In these experiments (reported in \cref{sec:exp_validation} and \cref{par: pe}) we used the test set of the standard version of \dataset{GSM8K}, \dataset{HellaSwag}, \dataset{ARC}, \dataset{Winogrande}, and \dataset{TruthfulQA}. Furthermore, we used 6 different models to test the different performance of the ability and performance estimator: \model{SOLAR-tail-10.7B-Merge-v1.0}, \model{FuseChat-7B-Slerp}, \model{NeuralPipe-7B-slerp}, \model{T3Q-Merge-Mistral7B}, \model{FuseChat-7B-TA}, and \model{supermario-slerp-v3}. These models were chosen as already available on the Open LLM Leaderboard.

%\subsection{Notation}

%In \cref{tab:notations} we outline a scheme of the notation used along the paper.

\begin{table}
\centering
\caption{Notation used in the paper.}
\label{tab:notations}
\vspace{10pt}
\resizebox{\columnwidth}{!}{%
\begin{tabular}{cl}
    \toprule
    \textbf{Notation} & \textbf{Description} \\ 
    \midrule
    $D$ & Full dataset. \\
    $\bar{D}$ & Reduced subset of the dataset. \\
    $D_{i}$ & Subdataset for task $i$. \\
    $\gamma_m$ & Latent abilities of model $m$. \\
    $\Gamma_m$ & True latent abilities of model $m$. \\
    $\gamma_m^{\{\mathrm{p,gp}\}-\mathrm{IRT}} $ &  Latent abilities of model $m$ via \pirt{} ability estimator.\\
    $\gamma_m^{\{\mathrm{mp,gmp}\}-\mathrm{IRT}} $ &  Latent abilities of model $m$ via \mpirt{} ability estimator.\\
    $\alpha_i, \beta_i$ & IRT parameters related to dataset item $i$. \\
    $\lambda$ & Interpolation coefficients for latent abilities. \\
    $\hat{\lambda},\hat{\gamma},\hat{\alpha},\hat{\beta}$ & MLE of the aforementioned parameters. \\
    $p_{i,m}$ & IRT model for datapoint $i$ and model $m$. \\
    \multirow{2}{*}{$\hat{p}_{i,m}$} & IRT model for datapoint $i$ and model $m$ \\
                    & parametrized by MLE estimators of $\alpha, \beta,\gamma, \lambda$. \\
    $\tilde{m}$ & Merged language model. \\
    $Y_{i,m}$ & Sample-level correctness of model $m$ for example $i$. \\
    $\hat{Z}^{\mpirt{}}$ & Merged performance estimator \mpirt{}. \\
    $\hat{Z}^{\gmpirt{}}$ & Generalized merged performance estimator \gmpirt{}. \\
    $F(m)$ & Fitness value of a model $m$. \\
    $\theta$ & Parameters being optimized in evolutionary search. \\
    $P_{\bar{F}_D}$ & Pareto front defined by function's set $\bar{F}$ and data $D$ \\
    $\theta^{\star}$ & Global optimum on $D$. \\
    $\hat{\theta}$ & Global optimum on $\bar{D}$. \\
    $N$ & Number of samples in the dataset. \\
    \bottomrule
\end{tabular}
}
\end{table}

%% file: B_Additional_Experiments/content.tex
\label{app:add-exp}
We report here additional experiments and analyses.

\subsection{Extract Step} \label{app:add-exp:extract-ste} 
In the extract step outlined in \cref{sec:estimate}, random sampling has been proposed as the main method to subsample the dataset $\bar{D} \subset D$. 
While we explored various dataset subsampling strategies, we ultimately opted for uniform random sampling, as our experiments showed that more complex approaches offered no significant advantage over this simpler method. In this section we report some of the experiments behind this decision and the two alternative methods tried in the extraction step: IRT Clustering (IRT), introduced by \citet{tinybenchmarks}, and a custom Representation Clustering (RC) method.

\subsubsection{IRT Clustering} Given a dataset $D$ and the parameter of a fitted IRT model $\alpha$ and $\beta$, one can define a low-dimensional embedding of each datapoint $i \in D$ by $E_i = [\alpha_i \| \beta_i]$. Therefore, IRT-clustering obtains a representative subset by first obtaining a clustering over this embedding space through $K$-Means, and then choosing the points closest to the centroids as representative samples.

\subsubsection{Representation Clustering} 

Let \(\{m_j\}_{j=1}^M\) be the set of endpoint models, and let \(D = \{x_i\}_{i=1}^N\) be our full dataset. For each sample \(x_i\), we first encode it into a high-dimensional vector by concatenating model-specific embeddings. Concretely, we compute:
\[
    E_{i,j} = \frac{1}{T_i}\sum_{t=1}^{T_i} E_{i,j,t} \in \mathbb{R}^d,
\]
where \(E_{i,j,t}\) is the embedding of the \(t\)-th token of sample \(x_i\) under model \(m_j\), and \(T_i\) is the number of tokens in \(x_i\). We form the concatenated representation:
\[
    E_i = [E_{i,1}\|E_{i,2}\|\cdots\|E_{i,M}] \in \mathbb{R}^{M\cdot d}.
\]
Since \(E_i\) can be very high-dimensional, we apply Principal Component Analysis (PCA) to project \(E_i\) onto a lower-dimensional space:
\[
    \tilde{E}_i = \text{PCA}_k(E_i) \in \mathbb{R}^k, \quad k \ll M\cdot d.
\]
Next, we apply \(k\)-means clustering to the reduced embeddings \(\{\tilde{E}_i\}_{i=1}^N\):
\[
    \min_{\{\mathbf{c}_k\}_{k=1}^K}\sum_{i=1}^N \min_{1\leq k \leq K} \|\tilde{E}_i - \mathbf{c}_k\|^2,
\]
where \(\mathbf{c}_k\) is the centroid of the \(k\)-th cluster. This partitions the dataset into \(K\) clusters, each capturing a distinct region of the representation space.
From each cluster \(k\), we select the representative sample \(x_{i_k^\star}\) whose embedding \(\tilde{E}_{i_k^\star}\) is closest to the centroid \(\mathbf{c}_k\):
\[
    i_k^\star = \arg\min_{x_i \in C_k} \|\tilde{E}_i - \mathbf{c}_k\|,
\]
where \(C_k\) is the set of samples assigned to cluster \(k\).
To approximate the full-dataset metrics from the selected subset \(\bar{D}=\{x_{i_k^\star}\}_{k=1}^K\), we assign a weight to each representative sample. Since the size of the cluster \(C_k\) indicates how prevalent that region of representation space is, we define $w_{i_k^\star} = \frac{|C_k|}{|D|}$.
These weights ensure that the contribution of each representative sample to the overall metric reflects the true proportion of samples that it represents in the original dataset. By evaluating a new model \(m\) only on \(\bar{D}\) and using \(\{w_{i_k^\star}\}\) to calculate a weighted average, we approximate \(m\)’s performance on the full dataset \(D\) at a fraction of the computational cost.

A schematic overview of the full process is outlined in \cref{alg:repr-clust}. 

\begin{algorithm}
\caption{Representation Clustering Extractor}\label{alg:repr-clust}
\begin{algorithmic}[1]
    \REQUIRE Dataset $D$, Endpoint Models $m_1,...,m_n$, Desired subset size $K$
    \ENSURE Subset of size $K$ with weights $w_i$

    \FOR{$i$ in $D$}
        \STATE $E_i \gets []$ 
        \FOR{$m$ in $\{m_1, \dots, m_n\}$}
        \STATE $E_{im} \gets$ embed $i$ with model $m$
        \STATE  $E_i \gets E_i | E_{im}$ 
        \ENDFOR
    \ENDFOR
    \STATE $\{\mathbf{E}_i\}_{i \in D} \gets \operatorname{PCA}(\{\mathbf{E}_i\}_{i \in D})$
    \STATE Apply k-means clustering to $\{\mathbf{E}_i\}_{i \in D}$, obtaining $K$ centroids $\{\mathbf{c}_k\}_{k=1}^K$
    \STATE For each cluster $k$, select the closest example $i_k^{\star} = \arg\min_{i \in D} \|\mathbf{E}_i - \mathbf{c}_k\|_2$
    \STATE Let $C_k = \{i \in D \ | \ \arg\min_{c \in \{\mathbf{c}_k\}_{k=1}^K} \|\mathbf{E}_i - c\|_2 = \mathbf{c}_k\}$ be the set of examples in cluster $k$
    \STATE Assign weights $w_{i_k^{\star}} = \frac{|C_k|}{|D|}$ for $k=1,...,K$
    \STATE \textbf{return}  $\left\{ i_k^{\star}, w_{i_k^{\star}} \right\}_{k=1}^{K}$
\end{algorithmic}
\end{algorithm}

\subsubsection{Experiments}
\begin{figure}
    \includegraphics[width=\linewidth]{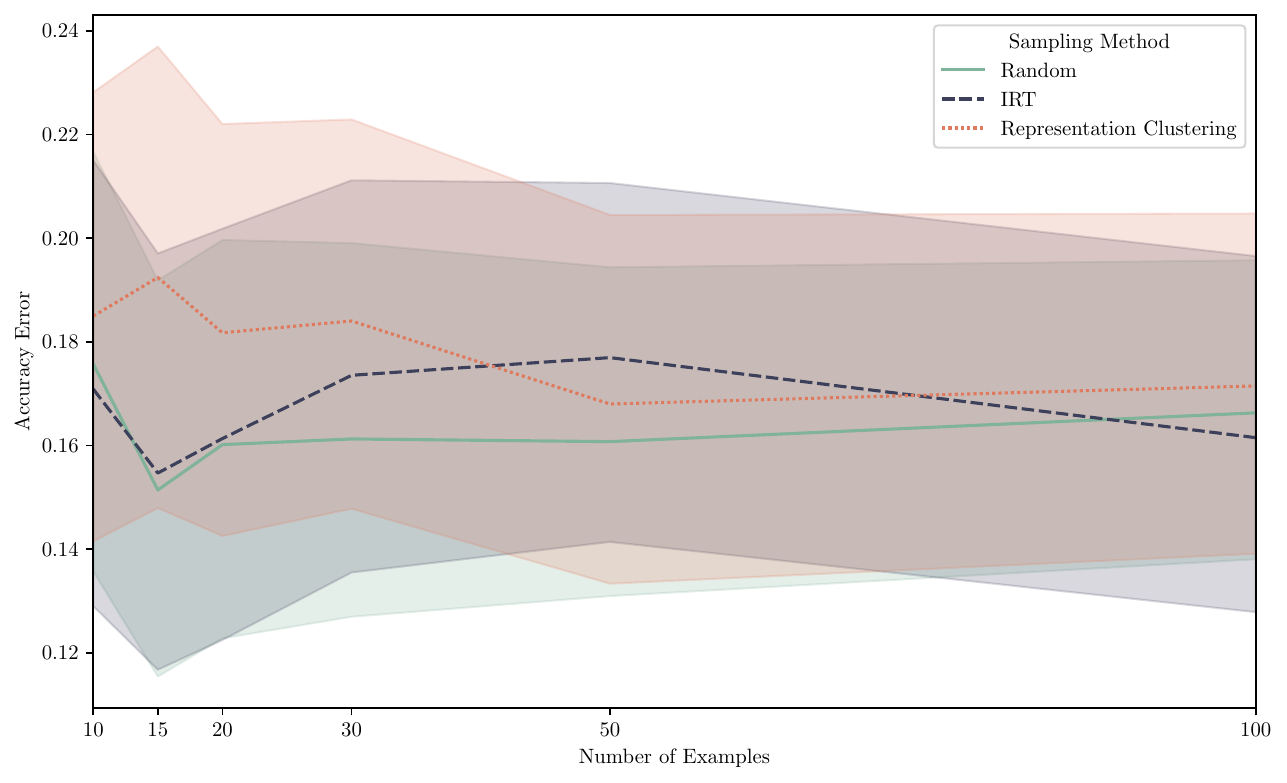}
    \caption{\textit{Extractors across Languages:} Absolute error of the estimated accuracy of Sample Extractors, averaged across merges of language-specific and English Math finetunings of \model{Mistral-7B-v0.1}, evaluated on translations of \dataset{GSM8K} and presented as a function of the number of dataset samples.}
    \label{fig:extractor_language}
\end{figure}

\begin{figure}
    \includegraphics[width=\linewidth]{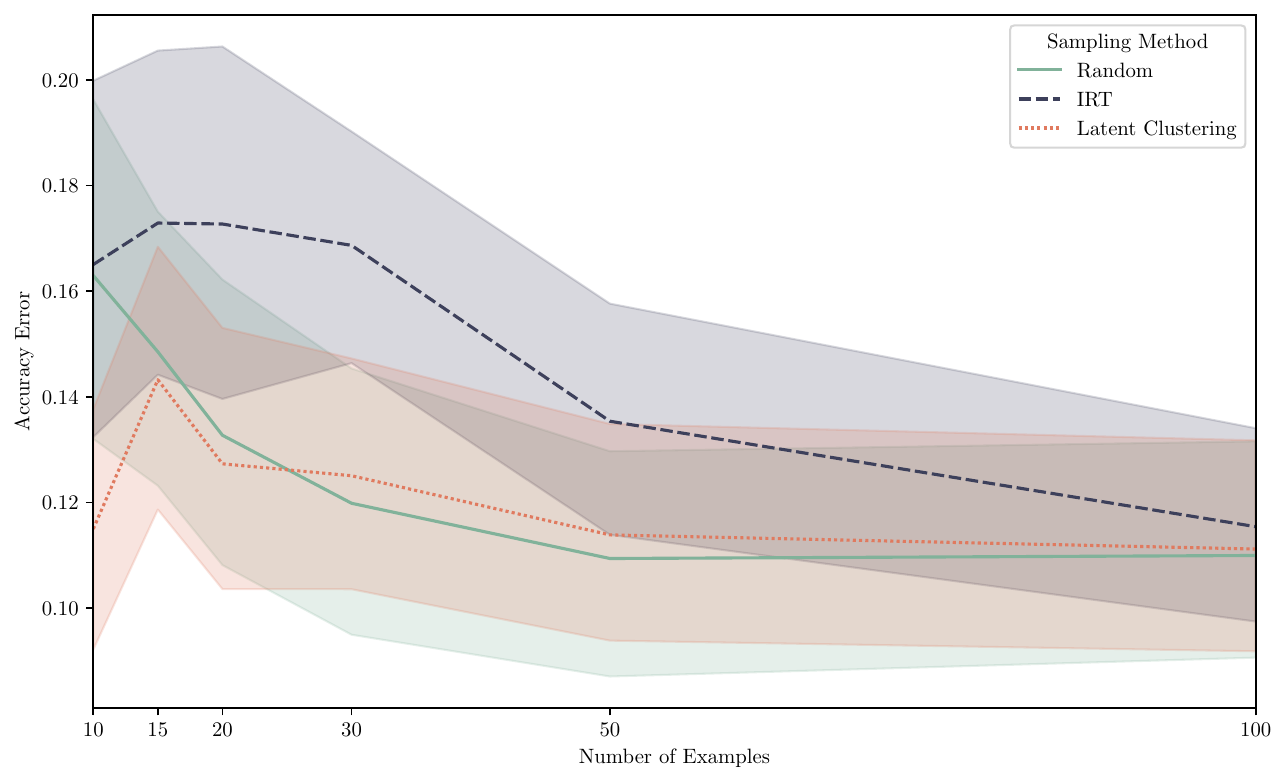}
    \caption{\textit{Extractors across Merges:} Absolute error of the estimated accuracy of Sample Extractors, averaged across merges of English Math models based on \model{Mistral-7B-v0.1}, evaluated on \dataset{GSM8K} and presented as a function of the dataset sample size.}
    \label{fig:extractor_merges}
\end{figure}

To compare the performance of the Sample Extractors, we followed a procedure similar to that described in \cref{par: pe}, computing the absolute estimation error for each extractor. For random sampling, the accuracy estimator was obtained via uniform averaging, whereas for IRT and RC it was obtained via weighted averaging. We evaluated the estimator in two different settings: (1) merging a math model with a language-tuned model (similar to the cross-lingual setting of \cref{sec: cross-lingual}) for several languages (Italian, German, Romanian, Dutch) and testing the extractor on the corresponding translations of \dataset{GSM8K} (see \cref{fig:extractor_language}), and (2) merging several math models and testing the extractor on the English version of \dataset{GSM8K} (see \cref{fig:extractor_merges}).

Focusing on \cref{fig:extractor_language}, we see that performance variability is somewhat higher (larger error bars) due to different language-specific datasets. Even so, Random sampling never falls behind IRT or RC, especially for small sample sizes. By the time the subset size reaches 50 or more examples, all three methods converge to comparable accuracy-error levels, underscoring the robustness of Random sampling. Instead, in \cref{fig:extractor_merges}, the trends are broadly similar for RC and Random sampling, while slightly worse for IRT.  Again, as the dataset sample size grows, overall error drops and the gap among methods narrows. 

To sum up, the Random sampler can sometimes lag slightly behind the more sophisticated IRT and RC. Nevertheless, neither of these methods has a clear advantage over the others. Given its simplicity and negligible overhead, the Random strategy stands out as a highly practical choice for dataset subsampling—especially when the marginal improvements of more complex methods do not clearly justify their added complexity.

\subsection{Estimation step} \label{app:estimation-step}

\subsubsection{Additional Experiment for Ability Estimator}
\begin{figure}
    \includegraphics[width=\linewidth]{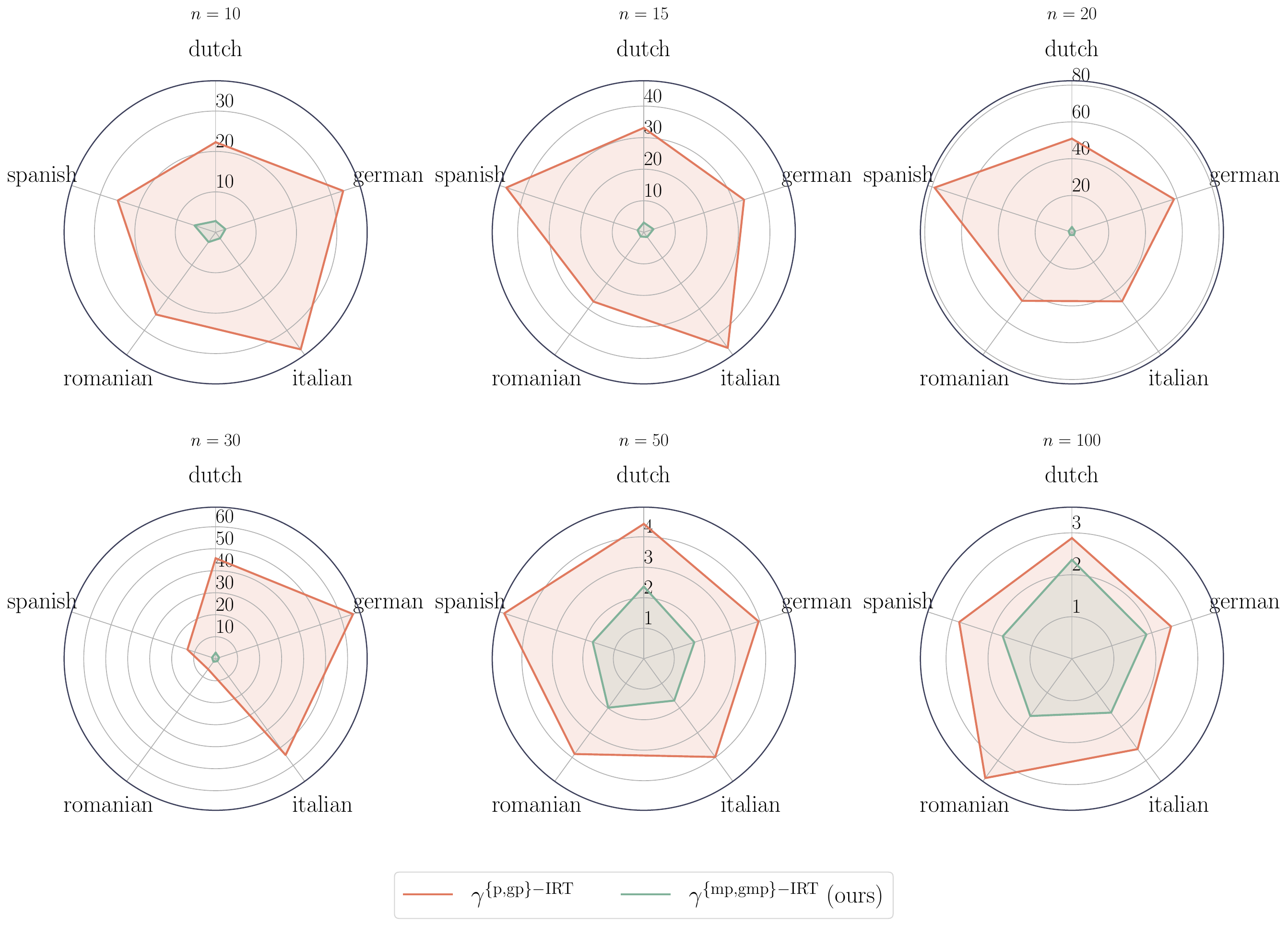}
    \caption{\textit{Ability Estimator over languages:} Euclidean distance (lower is better) between estimated and true abilities for different languages.}
    \label{fig:ability-estimator-comparison-all}
\end{figure}

\begin{figure}
    \includegraphics[width=\linewidth]{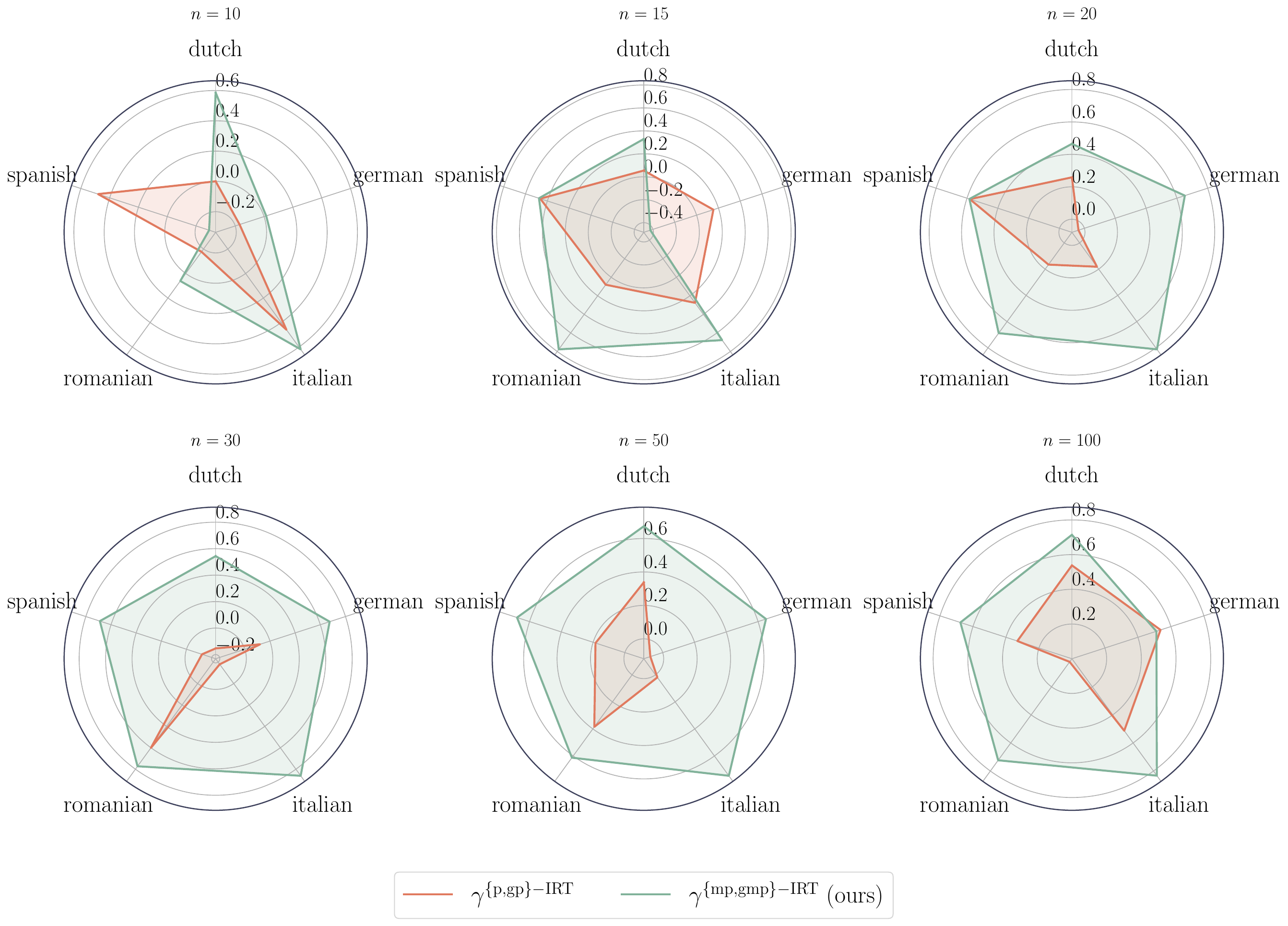}
    \caption{\textit{Ability Estimator over languages:} Cosine similarity (higher is better) between estimated and true abilities for different languages.}
    \label{fig:ability-estimator-comparison-cosine}
\end{figure}

We report in \cref{fig:ability-estimator-comparison-all} the Euclidean distance between the estimated and ground-truth ability vectors across different sample sizes. The results are consistent with the case $n=10, 20$ seen in \cref{fig:ability-estimator-comparison}, with our estimated ability vector being significantly closer to the ground-truth one compared to the ability vector estimated by pIRT and gp-IRT. Similarly, we report the corresponding cosine similarity in \cref{fig:ability-estimator-comparison-cosine}, confirming much higher similarity in our case.  
\begin{figure}
    \includegraphics[width=\linewidth]{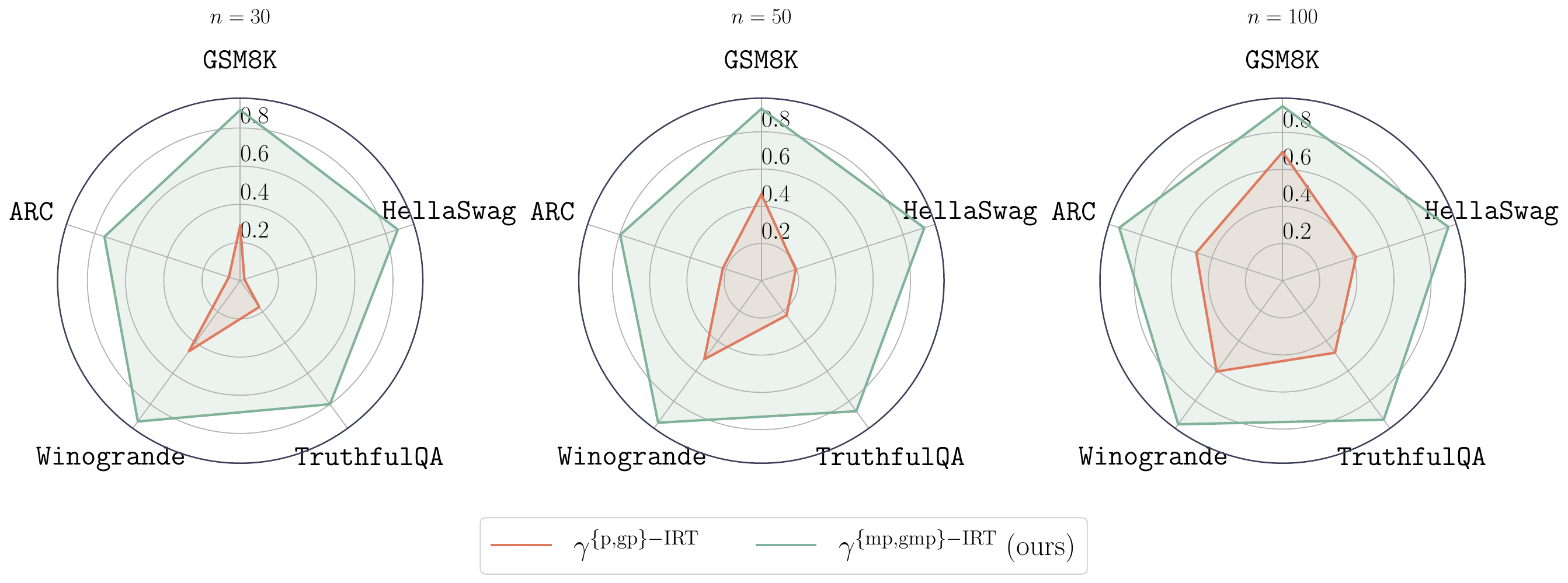}
    \caption{\textit{Ability Estimator over tasks:} Cosine similarity (higher is better) between estimated and true abilities for different tasks.}
    \label{fig:ability-estimator-comparison-cosine-tasks}
\end{figure}
\subsubsection{Additional Experiment for Performance Estimator} 
\begin{figure}
    \centering
    \includegraphics[width=\linewidth]{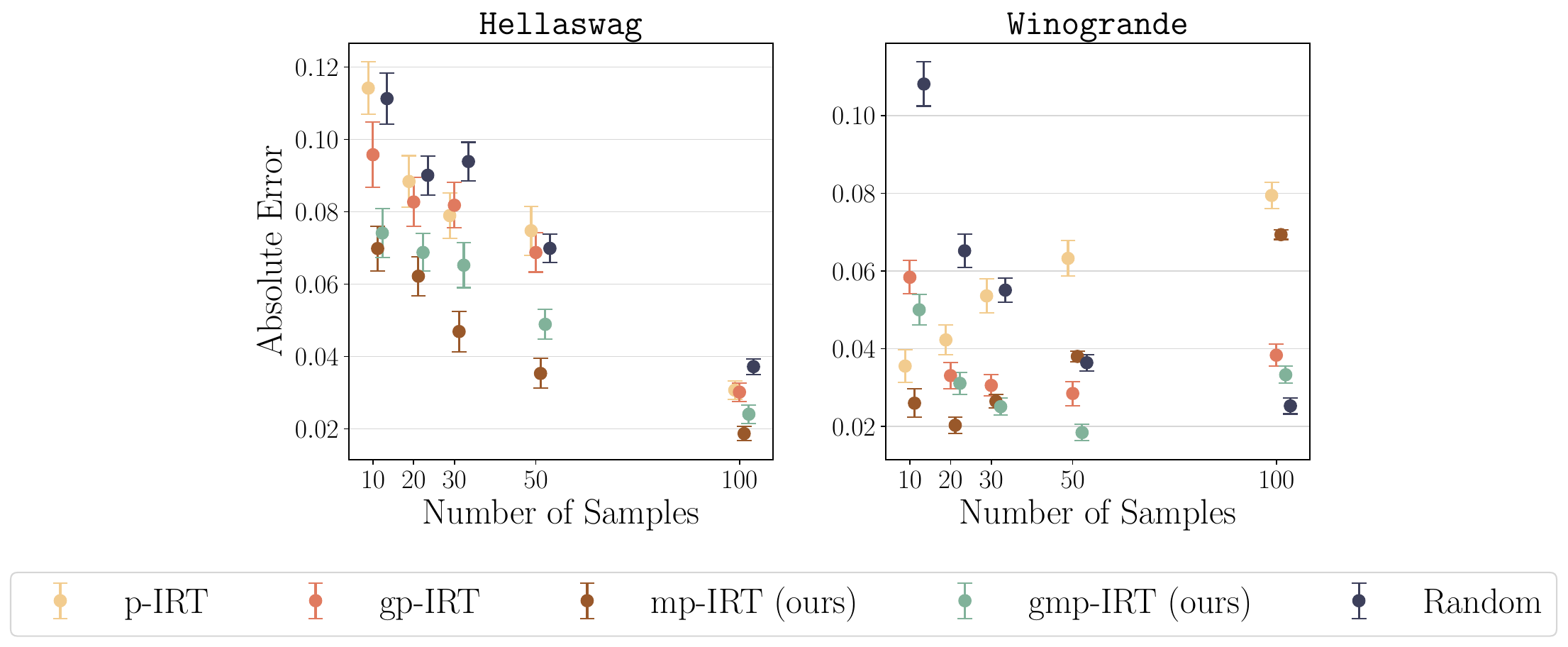}
    \caption{\textit{Performance Estimators over Winogrande and Hellaswag.} Absolute error of various estimators as a function of sample size (lower is better). gmp-IRT consistently achieves lower error.}
\label{fig:estimation_comparison_winogrande_hellaswag}
\end{figure}

We report in \cref{fig:estimation_comparison_winogrande_hellaswag} the evaluation of performance estimators across \dataset{Winogrande} and \dataset{Hellaswag}, extending the results in \cref{fig:estimation_comparison}.

\subsection{Evolve Step} 

\subsubsection{Additional Experiment for Multilingual Evolution: Self-Merging}
\label{app:add-exp-self-merging}

In this section, we present an ablation study to rigorously test whether the observed improvements in the merged models arise from genuine cross-lingual knowledge transfer or merely from fitting to the prompt template. To structure this analysis, we formalize our inquiry through two statistical hypotheses:

\paragraph{Null Hypothesis ($H_0$).} The improvements seen in the merged models are due to the model fitting itself on the prompt template, rather than any cross-lingual knowledge exchange.

\paragraph{Alternative Hypothesis ($H_1$).} The improvements arise from actual cross-lingual knowledge transfer---specifically, from the mathematical model to the linguistic model---and are not merely the result of fitting the prompt template.

To evaluate these hypotheses, we propose a \emph{self-merging} procedure. Concretely, we take the linguistic model and merge it with \emph{itself} using the standard \approach methodology outlined in \cref{alg:merge3}. Under $H_0$, if the improvements are solely due to the prompt template, merging the model with itself should lead to performance gains (i.e., the merged model would still ``fit'' the template). Conversely, under $H_1$, if cross-lingual knowledge transfer is responsible for the enhanced performance, self-merging should \emph{not} yield improvements. In fact, additional noise could even degrade performance relative to the baseline.

\begin{figure}
    \centering
    \includegraphics[width=.8\linewidth]{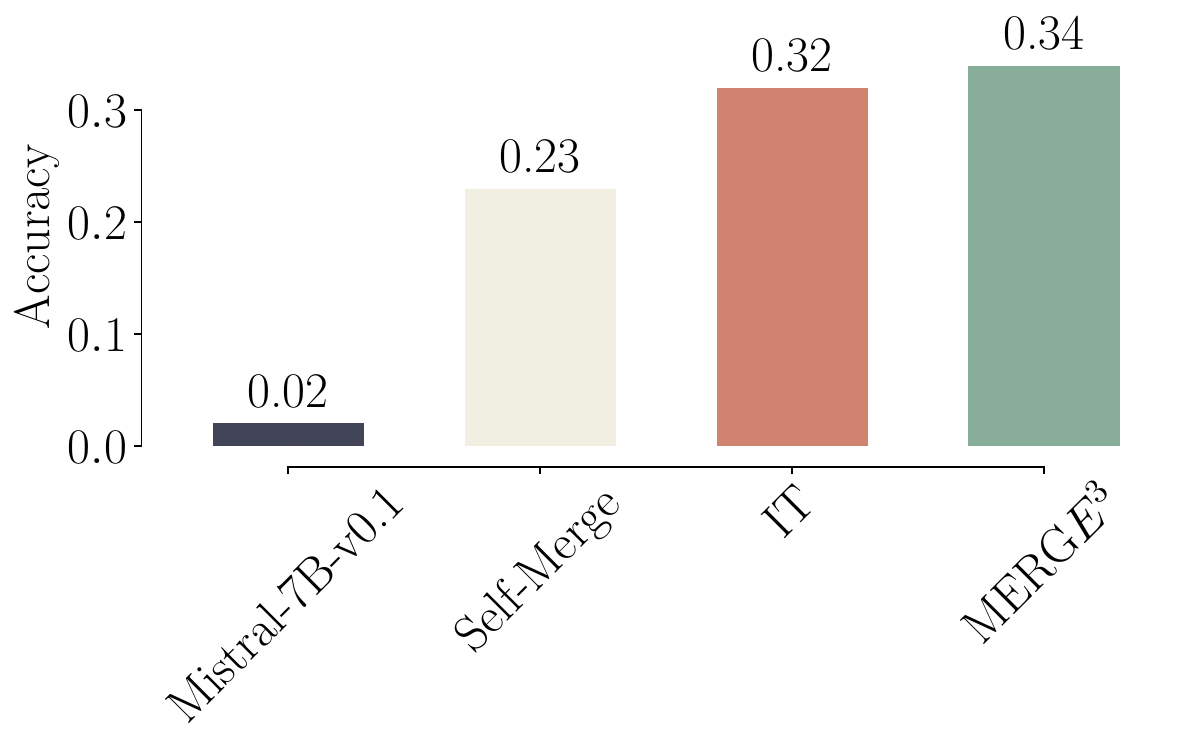}
    \caption{Accuracy of the base model (Mistral-7B), the Italian Endpoint (\texttt{IT}), Self-Merge and \approach models on the Italian-translated version of \dataset{GSM8k}.}
    \label{fig:self-merge}
\end{figure}

We conducted this self-merging experiment on the Italian model using the \dataset{GSM8K} dataset. The results, shown in \cref{fig:self-merge}, reveal that performance actually \emph{decreases} when the model is merged with itself. This observation strongly supports the alternative hypothesis ($H_1$): the performance gains in cross-lingual merges indeed stem from genuine knowledge transfer, rather than mere adaptation to a prompt template.

\subsubsection{Additional Experiment: Time \& FLOPs Requirements Evolutionary Merging}
\label{app:add-exp-4090}
\paragraph{Hardware Setting.} To compare the efficiency of different model evaluation strategies, we measured the time required to evolve merged LLM models using a single NVIDIA 4090 with 24GB of VRAM, and report the Throughput $R$ in \cref{tab:throughput_comparison}. 

\paragraph{Results Discussion.} Over a 12-hour period, we were able to evaluate 8 models on 1000 samples of \dataset{GSM8K}, allowing us to estimate that evaluating 1000 models would take approximately 62 days under similar conditions. In contrast, \approach enabled the evaluation of a larger number of merged models in significantly less time by using a reduced dataset. These results suggest that researchers and practitioners could leverage consumer-grade GPUs for efficient LLM merging and evaluation, making rapid experimentation of model merging methods more accessible. We report in \cref{tab:evolution_comparison} the estimated total time of the Evolve runs, which we calculated using the following formula:
\[
T(N_{\text{models}}) \;=\; \frac{N_{\text{models}}}{R_{\text{Dataset Size}}}
\]
\begin{table}[h]
    \centering
    \caption{Throughput (\( R \)) in models per hour for different sample sizes per fitness evaluation on \dataset{GSM8K}. These estimates are based on 12-hour Evolve runs on a single NVIDIA 4090 with 24GB of VRAM.}
    \label{tab:throughput_comparison}
    \vspace{10pt}
    \resizebox{\linewidth}{!}{
    \begin{tabular}{lccccc}
        \toprule
        \textbf{Sample size} & 1000 & 100 & 50 & 30 & 20 \\
        \midrule
        \textbf{Throughput (Models/Hour)} & 0.67 & 8.33 & 14.17 & 16.67 & 17.08 \\
        \bottomrule
    \end{tabular}
    }
\end{table}

\begin{table}[h]
    \centering
    \caption{Comparison of Evolve methods by number of trials, estimated total time on a single NVIDIA 4090, sample size used for Fitness computation, and final accuracy on \dataset{GSM8K}. The number of trials is the result of $\text{population size} \times \text{iterations}$, parameters of the Genetic Algorithms of each method, and represents the total number of merged models evaluated during the entire Evolve run.}
    \label{tab:evolution_comparison}
    \vspace{10pt}
    \resizebox{\linewidth}{!}{
    \begin{tabular}{lcccc}
        \toprule
        \textbf{Method} & \textbf{$N_{\text{models}}$} & \textbf{Estimated total time} & \textbf{Sample size} & \textbf{Accuracy} \\
        \midrule
        EvoLLM-JP-7B & 1000 & 62 days & 1000 & 0.49 \\
        MERGE$^3_{100}$ & 175 & 21h & 100 & 0.42 \\
        MERGE$^3_{50}$ & 175 & 12h 20m & 50 & 0.38 \\
        MERGE$^3_{30}$ & 175 & 10h 30m & 30 & 0.38 \\
        MERGE$^3_{20}$ & 175 & 10h 15m & 20 & 0.34 \\
        \bottomrule
    \end{tabular}
    }
\end{table}

\paragraph{FLOPs Calculation.} We provide a Jupyter Notebook that describes the FLOPs calculations for our experiments in the supplementary material, based on the \textit{calc-flops} library\footnote{ \url{https://github.com/MrYxJ/calculate-flops.pytorch}.}. This script has been used to estimate the FLOPs for the experiment in \Cref{fig:speed_accuracy}.

%% file: D_Proofs/content.tex
We outline in \cref{tab:notations} a scheme of the notation used throughout the paper.

\subsection{Proof of \Cref{thm:eps-opt-preserve}}
\label{proof:thm-eps-opt-preserve}

\begin{proof}
Let \(m := F(\theta^*;D)\) and \(\hat{m} := F(\hat{\theta};\bar{D})\).  
We must show that \(\lvert\,m - \hat{m}\rvert \le \epsilon\).

\begin{enumerate}
\item 
By \(\epsilon\)-stability, for \emph{all} \(\theta\in\Theta\):
\[
  \bigl\lvert F(\theta;D) \;-\; F(\theta;\bar{D})\bigr\rvert
  \;\le\;
  \epsilon.
\]
In particular, for \(\theta = \theta^*\),
\[
  \bigl\lvert F(\theta^*;D) \;-\; F(\theta^*;\bar{D})\bigr\rvert
  \;\le\;
  \epsilon.
\]
Hence
\[
  F(\theta^*;\bar{D}) 
  \;\ge\;
  F(\theta^*;D)\;-\;\epsilon
\]
and
\[
  F(\theta^*;\bar{D}) 
  \;\le\;
  F(\theta^*;D)\;+\;\epsilon.
\]

\item 
Since \(\hat{\theta}\) is the minimizer of \(F(\cdot;\bar{D})\), we have
\[
  F(\hat{\theta};\bar{D}) 
  \;\le\; 
  F(\theta^*;\bar{D}).
\]
Because \(\theta^*\) is the minimizer of \(F(\cdot;D)\), 
\[
  F(\hat{\theta};D) 
  \;\ge\; 
  F(\theta^*;D).
\]

\item 
To bound \(\hat{m}-m\), we can add and subtract $F(\theta^*;\bar{D})$ to have
\begin{align*}
  \hat{m}-m
  &\;=\;
  \Bigl(F(\hat{\theta};\bar{D}) \;-\; F(\theta^*;\bar{D})\Bigr) \\
  &\;+\;
  \Bigl(F(\theta^*;\bar{D}) \;-\; F(\theta^*;D)\Bigr).
\end{align*}
The first term is \(\le 0\) (since \(\hat{\theta}\) is a minimizer on \(\bar{D}\)), 
and the second term is \(\le \epsilon\).  
Hence
\[
  \hat{m}-m
  \;\le\;
  0 + \epsilon
  \;=\;
  \epsilon.
\]

\item 
Analogously, to bound \(m-\hat{m}\), we can rewrite
\begin{align*}
  m-\hat{m}
  &\;=\;
  \Bigl(F(\theta^*;D) \;-\; F(\hat{\theta};D)\Bigr) \\
  &\;+\;
  \Bigl(F(\hat{\theta};D) \;-\; F(\hat{\theta};\bar{D})\Bigr).
\end{align*}
The first term is \(\le 0\) (since \(\theta^*\) is a minimizer on \(D\)), 
and the second term is \(\le \epsilon\).  
Thus,
\[
  m-\hat{m}
  \;\le\;
  0 + \epsilon
  \;=\;
  \epsilon.
\]

\item 
Combining these inequalities:
\[
  -\epsilon 
  \;\le\; 
  \hat{m}-m 
  \;\le\; 
  \epsilon
  \quad\Longrightarrow\quad
  \lvert
    m - \hat{m}
  \rvert
  \;\le\;
  \epsilon.
\]
Hence 
\(\bigl\lvert F(\theta^*;D) - F(\hat{\theta};\bar{D})\bigr\rvert\le\epsilon\), 
completing the proof.
\end{enumerate}
\end{proof}

\subsection{Proof of \Cref{thm:expected-eps-stability}}
\label{proof:thm-expected-eps-stability}
\begin{proof}
By hypothesis, for every \(\theta\in\Theta\),
\[
  \mathbb{E}_{\bar{D}}\!\Bigl[
    \bigl\lvert F(\theta;D) \;-\; F(\theta;\bar{D})\bigr\rvert
  \Bigr]
  \;\le\;\epsilon.
\]
Using Jensen's inequality for the absolute value,
\begin{align*}
  &\bigl\lvert
    \mathbb{E}_{\bar{D}}\!\bigl[
      F(\theta;D) \;-\; F(\theta;\bar{D})
    \bigr]
  \bigr\rvert \\
  &\;\le\;
  \mathbb{E}_{\bar{D}}\!\Bigl[
    \bigl\lvert
      F(\theta;D) \;-\; F(\theta;\bar{D})
    \bigr\rvert
  \Bigr] 
  \;\le\;\epsilon.
\end{align*}
Hence,
\[
  -\epsilon 
  \;\le\;
  \mathbb{E}_{\bar{D}}\!\bigl[
    F(\theta;\bar{D}) - F(\theta;D)
  \bigr]
  \;\le\;
  \epsilon
\]
for each fixed $\theta$. It thus follows that
\[
  \mathbb{E}_{\bar{D}}\!\bigl[F(\theta;\bar{D})\bigr]
  \;\le\;
  F(\theta;D) + \epsilon
\]
and
\[
  \mathbb{E}_{\bar{D}}\!\bigl[F(\theta;\bar{D})\bigr]
  \;\ge\;
  F(\theta;D) - \epsilon.
\]
Consequently,
\[
  \min_{\theta\in\Theta}\,
  \mathbb{E}_{\bar{D}}\!\bigl[F(\theta;\bar{D})\bigr]
  \;\le\;
  \min_{\theta\in\Theta}
  \bigl[
    F(\theta;D) + \epsilon
  \bigr]
  \;=\;
  m^* + \epsilon,
\]
where \(m^* := \min_{\theta\in\Theta} F(\theta;D)\).  
Meanwhile, by a min-versus-expectation (Jensen-type) inequality,
\[
  \mathbb{E}_{\bar{D}}\!\bigl[\min_{\theta\in\Theta}\,
  F(\theta;\bar{D})\bigr]
  \;\ge\;
  \min_{\theta\in\Theta}
  \mathbb{E}_{\bar{D}}\!\bigl[
    F(\theta;\bar{D})
  \bigr].
\]
Hence,
\begin{align*}
  \mathbb{E}_{\bar{D}}\!\bigl[\widehat{m}(\bar{D})\bigr]
  &\;=\;
  \mathbb{E}_{\bar{D}}\!\Bigl[\min_{\theta\in\Theta}
    F(\theta;\bar{D})
  \Bigr]
  \\ 
  &\;\ge\; 
  \min_{\theta\in\Theta}
  \mathbb{E}_{\bar{D}}\!\bigl[
    F(\theta;\bar{D})
  \bigr]
  \;\ge\;
  m^* - \epsilon.
\end{align*}
Combining these two bounds results in
\[
  m^* - \epsilon
  \;\le\;
  \mathbb{E}_{\bar{D}}\!\bigl[\widehat{m}(\bar{D})\bigr]
  \;\le\;
  m^* + \epsilon
\]
and, therefore,
\[
\Bigl|
    m^*
    \;-\;
    \mathbb{E}_{\bar{D}}\!\bigl[\widehat{m}(\bar{D})\bigr]
  \Bigr|
  \;\le\;\epsilon.
\]
\end{proof}

\subsection{Proof of \Cref{prop:estimator-unbiased}} \label{proof:prop-estimator-unbiased}
\begin{proof}
We must show that 
\[
\bigl\lvert
  \mathbb{E}\!\bigl[\hat{Z}_{jl} \,\bigm|\,
    Y_{i_0l},\dots,Y_{i_kl}
  \bigr]
  \;-\;
  \mathbb{E}\!\bigl[Z_{jl} \,\bigm|\,
    Y_{i_0l},\dots,Y_{i_kl}
  \bigr]
\bigr\rvert
\;\to\;0
\]
in probability as
$|\hat{I}|\to\infty$.
Under the assumptions of the proposition (including linear inheritance of abilities, 
\(\hat{\lambda}\to\lambda\) in probability, and bounded \(\|\alpha_i\|\)), 
we may bound this difference as follows:
\begin{align*}
&\bigl\lvert
  \mathbb{E}\!\bigl[\hat{Z}_{jl} \, \bigm|\,
    Y_{i_0l}, \ldots, Y_{i_kl}
  \bigr]
  \;-\;
  \mathbb{E}\!\bigl[Z_{jl} \,\bigm|\,
    Y_{i_0l}, \ldots, Y_{i_kl}
  \bigr]
\bigr\rvert\\
&\;\le\;
\frac{1 - \hat{\lambda}}{|I_j \setminus \hat{I}_j|}
\sum_{i \in I_j \setminus \hat{I}_j}
\Bigl|
  \sigma\!\bigl((\hat{\lambda}_1\theta_{l_1}
    +\hat{\lambda}_2\theta_{l_2})^\top \alpha_i 
    \;-\;\beta_i\bigr)\\
  &\;-\;
  \sigma\!\bigl(\theta_{l_m}^\top \alpha_i 
    \;-\;\beta_i\bigr)
\Bigr|.
\end{align*}
Since \(\sigma\) is \(1/4\)-Lipschitz on \(\mathbb{R}\), we have
\begin{align*}
    &= \;\le\;
    \frac{1}{|I_j|}
    \sum_{i \in \hat{I}_j}
    \Bigl|
      \bigl(
        (\hat{\lambda}_1\theta_{l_1}
         +\hat{\lambda}_2\theta_{l_2})
        \;-\;
        \theta_{l_m}
      \bigr)^\top \alpha_i
    \Bigr| \\ 
    &\;\le\;
    \frac{1}{|I_j|}
    \sum_{i \in \hat{I}_j}
    \|\alpha_i\|_2
    \,
    \|(\hat{\lambda}_1\theta_{l_1}
        +\hat{\lambda}_2\theta_{l_2})
      \;-\;\theta_{l_m}\|_2.
\end{align*}
Since \(\sup_{i\in I_j}\|\alpha_i\|_2 \le c\), it follows that
\[
\;\le\;
c\,
\bigl\|
  (\hat{\lambda}_1-\lambda_1)\,\theta_{l_1}
  \;+\;
  (\hat{\lambda}_2-\lambda_2)\,\theta_{l_2}
\bigr\|_2
\;\to\;0
\]
in probability as $|\hat{I}|\to\infty$.
(The last step uses \(\hat{\lambda}\to\lambda\) in probability, with 
\(\theta_{l_1}, \theta_{l_2}\) fixed in \(\mathbb{R}^d\).)  
Hence \(\hat{Z}_{jl}\) converges in probability to \(Z_{jl}\), completing the proof.
\end{proof}

\subsection{Proof of \Cref{thm:eps-opt-preserve}}
\label{proof:eps-opt-preserve}
\begin{proof}
By \Cref{prop:estimator-unbiased}, \(\hat{Z}^{\mathrm{mp\text{-}IRT}}\) 
becomes arbitrarily close (in probability) to \(Z\) as 
\(\lvert \bar{D}\rvert\to\infty\).  Under standard regularity conditions, this 
implies 
\[
  \bigl|
    Z(\theta;D)
    \;-\;
    \hat{Z}^{\mathrm{mp\text{-}IRT}}(\theta;\bar{D})
  \bigr|
  \;\le\;\epsilon
\]
in expectation, for all sufficiently large $\lvert \bar{D}\rvert$, hence \(\hat{Z}^{\mathrm{mp\text{-}IRT}}\) is \(\epsilon\)-stable in expectation.  
Applying \Cref{thm:expected-eps-stability} completes the argument.
\end{proof}